\documentclass{article}
\usepackage{arxiv}

\usepackage[utf8]{inputenc} 
\usepackage[T1]{fontenc}    
\usepackage{hyperref}       
\usepackage{url}            
\usepackage{booktabs}       
\usepackage{amsfonts}       
\usepackage{nicefrac}       
\usepackage{microtype}      
\usepackage{lipsum}

\usepackage[figuresright]{rotating}

\usepackage{amssymb}
\usepackage{amsthm}
\usepackage{amsmath}
\usepackage[normalem]{ulem}   

\usepackage[ruled,vlined,linesnumbered]{algorithm2e}
\usepackage{setspace}

\usepackage{booktabs}
\usepackage{multirow}

\usepackage{subfigure}
\newcommand{\etal}{\textit{et al.}}

\usepackage{array, longtable, tabularx}
\usepackage{pdflscape}

\newcommand{\minus}{\scalebox{0.75}[1.0]{$-$}}
\newcommand{\K}{\operatornamewithlimits{K}}
\newcommand{\Var}{\mathrm{Var}}

\usepackage{graphicx}
\usepackage[super]{nth}

\usepackage{xcolor}

\usepackage{lineno,hyperref}
\modulolinenumbers[5]

\title{Analytic Continued Fractions for Regression: \\A Memetic Algorithm Approach}

\author{
  Pablo Moscato\\
  School of Electrical Engineering and Computing,\\ The University of Newcastle,\\
Callaghan, NSW 2308, Australia\\
  \texttt{Pablo.Moscato@newcastle.edu.au} \\
   \And
 Haoyuan Sun \\
  California Institute of Technology,\\
  Pasadena, CA, USA\\
  \texttt{hsun2@caltech.edu} \\
   \AND
   Mohammad Nazmul Haque \\
   School of Electrical Engineering and Computing,\\ The University of Newcastle,\\
    Callaghan, NSW 2308, Australia\\
   \texttt{Mohammad.Haque@newcastle.edu.au} \\
}

\begin{document}
\maketitle

\begin{abstract}
We present an approach for regression problems that employs analytic continued fractions as a novel representation. Comparative computational results using a memetic algorithm are reported in this work. Our experiments included fifteen other different machine learning approaches including five genetic programming methods for symbolic regression and ten machine learning methods. The comparison on training and test generalization was performed using 94 datasets of the Penn State Machine Learning Benchmark. The statistical tests showed that the generalization results using analytic continued fractions provides a powerful and interesting new alternative in the quest for compact and interpretable mathematical models for artificial intelligence.  
\end{abstract}


\section{Introduction}

Symbolic regression is a unique type of multivariate regression analysis in which the goal is to find the mathematical expression of an unknown target function that would fit a dataset $S=\{(\mathbf{x^{(i)}}, y^{(i)})\}$, i.e. a set of pairs of an unknown multivariate target function $f : \mathbb{R}^n \to \mathbb{R}$. 
It has been argued that when analysing experimental data for decision making 
symbolic regression methods should at least be used to complement standard multivariate analysis~\cite{Duffy2002}.
Compared with the output of artificial neural network approaches, the models generated by symbolic regressions are generally more amenable to downstream studies via uncertainty propagation and sensitivity analysis and thus more ``explainable'' 
\cite{10.1007/978-3-642-32378-2_8,sun_ouyang_zhang_zhang_2019}. 
The other benefit of symbolic regression is the lack of assumptions on prior knowledge on the underlying process or mechanism which produced the observed data~\cite{DBLP:books/sp/19/MoscatoV19}.
This allows researchers to explore problem domains for which they have incomplete knowledge and identifying underlying trends and patterns without subjecting human bias.

Over the past three decades, symbolic regression had produced an impressive number of results in many applications.
For instance, symbolic regression has helped to extract physical laws using experimental data of chaotic dynamical systems without any knowledge of Newtonian mechanics~\cite{Schmidt2009}, which then motivated the data-driven discovery of hidden relationships in astronomy~\cite{Graham_2013}. 
More recent applications include prediction of friction systems performance~\cite{DBLP:conf/gecco/KronbergerKPN18}, 
identification of nonlinear relationships in fMRI data~\cite{DBLP:conf/eurogp/MartensKM17},
radiotherapy dose reconstruction of childhood cancer survivors~\cite{DBLP:conf/gecco/VirgolinABWB18}, also in the oncology field our own work on uncovering mechanisms of drug response in cancer cell lines using genomic and experimental data~\cite{Fitzsimmons_Moscato_2018}, 
predicting wind farm output from weather data~\cite{wind},
energy consumption forecasting~\cite{DBLP:conf/ipmu/DelgadoRCCJ18}, 
computer game scene generation~\cite{DBLP:journals/ijcgt/FradeVC09},
Boolean classification~\cite{DBLP:conf/eurogp/MuruzabalCF00}. 
They have also played a role in the elicitation of functional constructs from surveys~\cite{10.1371/journal.pone.0102768} and in the analysis of consumer and business data~\cite{Moscato_deVries}. 

One common approach to implementing symbolic regression is Evolutionary Computation (EC).
EC is a family of optimization algorithms inspired by biological evolution, in particular, building upon Darwin's theory of natural selection.
In EC, a population of candidate solutions (of a problem, generally posed as an optimization one) is subject to a set of heuristics and exact algorithms to produce new solutions, while less desirable solutions are being removed from the population currently under consideration.

EC approaches to symbolic regression are commonly based on Genetic Programming (GP) with a tree-based representation. Karabioga~\etal\: proposed Artificial Bee Colony Programming which also used the tree-based representation method for symbolic regression in~\cite{karaboga2012artificial} and the method showed competitive performance against GP-based methods.
Each solution (aka a mathematical expression) is written as a syntax tree and new solutions are produced by exchanging subtrees of two solutions (crossover) or modifying a syntax element, such as a binary operator (mutation)~\cite{DBLP:books/sp/19/MoscatoV19}.
Although highly popular, several researchers noted that recombination methods based on sub-tree crossovers have shown not to be better than some simple mutation of the sub-branches~\cite{Clegg:2007}. 
Clegg~\etal~in~\cite{Clegg:2007} cite previous contributions to this issue by Angeline~\cite{angeline1997subtree} and Luke and Spector~\cite{luke1997comparison}, in which they stated that ``due to findings like these, some people now implement their GP's without using crossover at all, i.e. using mutation only.''

We believe that this problem difficulty in symbolic regression could be addressed with generic problem-domain information about function approximation to search for better models. Like GP, Memetic Algorithms (MAs) is a generic denomination for a population-based approach to solve complex problems. However, MA takes explicit advantage of heuristic and exact methods in which solutions are individually optimized and also recombined and changed to improve the diversity of the population~\cite{DBLP:series/sci/2012-379}. 
First started at the California Institute of Technology three decades
ago~\cite{pablo,DBLP:series/sci/Moscato12}, research in MAs has demonstrated over the past three decades that problem-domain information can be used to produce local search (LS) methods that can significantly accelerate the evolutionary process. 
Trujillo~\etal ~in~\cite{leonardo:hal-01388426} recognize this fact and they point that, in contrast, local search has been underused in Genetic Programming. In their view, some of the problems faced by Genetic Programming are linked to the use of a tree-based representation of solutions. 
They conclude \textit{``that numerical LS and memetic search is seldom integrated in most GP systems''} and that \textit{``The fact that memetic approaches have not been fully explored in GP literature, opens up several areas (for) future lines of inquiry''}. 
We agree with this statement, in fact, of the 3918 publications we found about Memetic Algorithms (MAs) on the bibliographic database \textit{Web of Science} (on 20/11/2019), we have identified very few regarding the use of local search for symbolic regression. However, it is also true that some researchers have been trying to address the need of including individual optimization to existing Genetic Programming approaches, e.g.
\cite{ISI:000232173100154}, \cite{ISI:000335565200004}, \cite{DBLP:conf/gecco/FfranconS15}, \cite{DBLP:conf/iiaiaai/SemenkinaS15}. 
While this list is probably not comprehensive, it is recognized that introducing individual optimization steps into EA methods based on current representations for solutions has been a challenge for symbolic regression approaches.

In this paper, we introduce a new approach to regression with a memetic algorithm and we analyze its performance against other existing implementations of symbolic regression and machine learning approaches.
In particular, our contributions are as follows:
\begin{itemize}
    \item We introduce a novel method to represent mathematical expressions with analytic continued fractions by drawing inspirations from Pad\'{e} approximants.
    We discuss the advantages of this particular representation over the more traditional syntax-tree based representation.
    \item We implement a MA for symbolic regression with the continued fraction representation, a hierarchical population structure to manage the quality of the population of solutions, and an individual search method based on the Nelder-Mead algorithm.
    \item We compare our MA-based approach with 15 other state-of-the-art implementations of symbolic regression with 94 benchmark data sets. We demonstrate that our algorithm is able to extrapolate well-fitting relationships and its performance is comparable to other methods.
\end{itemize}

Following this introduction the remainder of this paper is organized as follows: the datasets and methods used in Section~\ref{data-and-methods}, in particular, the memetic algorithm is described in Section~\ref{sec:method}; we then present an illustrative one-dimensional example of using symbolic regression to approximate an important special function in mathematics, the Gamma Function in Section~\ref{sec:learning-the-gamma-function}. The computational results are presented in Section~\ref{sec:result}
followed by their discussion in Section~\ref{sec:discussion}. Finally, Section \ref{sec:conclude} contains concluding remarks and discusses the possibility of future work in the area.

%
%
\section{Data and Methods}
\label{data-and-methods}
In this section, we describe the proposed methods and datasets used to estimate its performance. We describe in detail the proposed symbolic regression method's representation, followed by the memetic algorithm for model identification, and an illustrative example of the proposed method in approximating the gamma function. Next, the section will describe the experimental procedures and datasets used to measure the performance of the proposed method.

\subsection{On representations and guiding functions}
\label{Representations-for-symbolic regression}
The paradigm of genetic programming (GP)~\cite{koza-gp} has been highly influential for the development of symbolic regression methods (e.g.~\cite{gptips}). 
In most cases, these GP approaches work by maintaining a population of models
that fit the data on the samples given as a training set, with ``mutations'' (i.e. small structural changes in the structures that code for these models) randomly changing the population. 
In addition, some ``recombination'' methods are used; generally these are simple heuristics involving the structures exchanging significant parts of them to create new individuals.  These processes generate new structures and, consequently, new models and a ``guiding function'' is used to somehow bias the search for better models, by encouraging the better models to be used more frequently in future iteration of the evolutionary process. 
In the area of symbolic regression, the ``guiding function'' is often some sort of merit function, e.g. the \textit{Mean Squared Error} (MSE) or the \textit{Mean Absolute Difference} (MAD). 
Using a plethora of different techniques, in general some part of the population is eliminated and some of the new models obtained via mutation and recombination may remain in the population (while less performing models according to the guiding function are deleted). Some other approaches to select the new population of solutions involve other considerations like data preprocessing~\cite{block} and the use of metrics that quantify the population diversity to avoid a premature convergence of the population to the same type of model or very similar variants on the same theme~\cite{fastsr}.

\subsubsection{Representation of models by parsing trees}
The syntax tree representation is simple and intuitive, and frequently used in evolutionary computation. Due to the influence of tree-based techniques for other problem domains, symbolic regression inherited this representation in GPs. 
Mathematical expressions were created by parsing tree structures and the evolutionary operators of mutation and recombination were defined on them. 
In a syntax tree representation, each node represents a mathematical operator, a variable, or a constant; and for each node representing an operator, its arguments are given as the children of this node. This approach has become very popular and the open-source GPTIPS~\cite{gptips} and the commercial software Eureqa~\cite{Schmidt2009, Eureqa:Software} are just two good examples of the many implementations that use this technique.

These implementations generally required user-defined \textit{``building blocks''}, the operands or functions that normally include the four arithmetic functions (i.e. addition, subtraction, multiplication and division), but other mathematical functions like logarithmic, exponentiation, trigonometric functions, etc., are sometimes needed as per implicit requirement of the problem characteristics. 
This breaks the assumption that we are not using problem domain information (i.e. we may be biasing the GP to bring solutions that involve some sinusoidal component for model building, by including the function sin(), while it may not be required at all). In addition, a representation based on parsing syntax trees, even with only the four basic arithmetic functions, could be handicapped to uncover some particular types of models that frequently are the best explanation in many real world problem. We refer to those models that can be defined as the ratio of two polynomials on the variables. The syntax tree representation has trouble uncovering these models in the search space. This is particularly evident after the population has evolved for some generations and most of the existing polynomial models are fitting the data relatively well, however, in general, it is unlikely that from those, via the use of division, we can produce a ratio of two polynomials that also fit the data well. 

The subtle way by which a powerful representation can create an implicit bias is generally not discussed in the literature. Correctness and completeness, instead of evolvability, seems to be the main concern, and functions defined as the ratio of higher order polynomials in subsets of the set of domain variables are less likely to be generated during the evolutionary computation run. 
Consequently, since both recombination and mutation fail to create new models that are competitive to remain in the population, we reach a stagnation of the search process. As models become increasingly similar, we reach premature convergence, which limits the performance of GP-based approaches in many symbolic regression applications. 

\subsubsection{Representation via multivariate Pad\'e approximants}
One possible way to address this issue would be to find a representation that restricts all models to be rational functions (i.e. ratios of two polynomials). We first recall the following important definitions. An \textit{analytic} function is defined as any function that can be written as a convergent power series in a neighborhood of each point in its domain $D$. A \textit{holomorphic} function is a complex-valued function of one or more complex variables that is, at every point of its domain, complex differentiable in a neighborhood of the point. 
A \textit{meromorphic} function on an open subset $D$ is a function that is holomorphic on all of $D$ except for a discrete set of isolated points, called the \textit{poles} of the function. Every meromorphic function on $D$ can be expressed as the ratio between two holomorphic functions defined on $D$ (with the denominator not being the constant 0), and any pole must coincide with a zero of the denominator. 

\begin{figure}
    \centering
    \includegraphics[width=0.8\columnwidth]{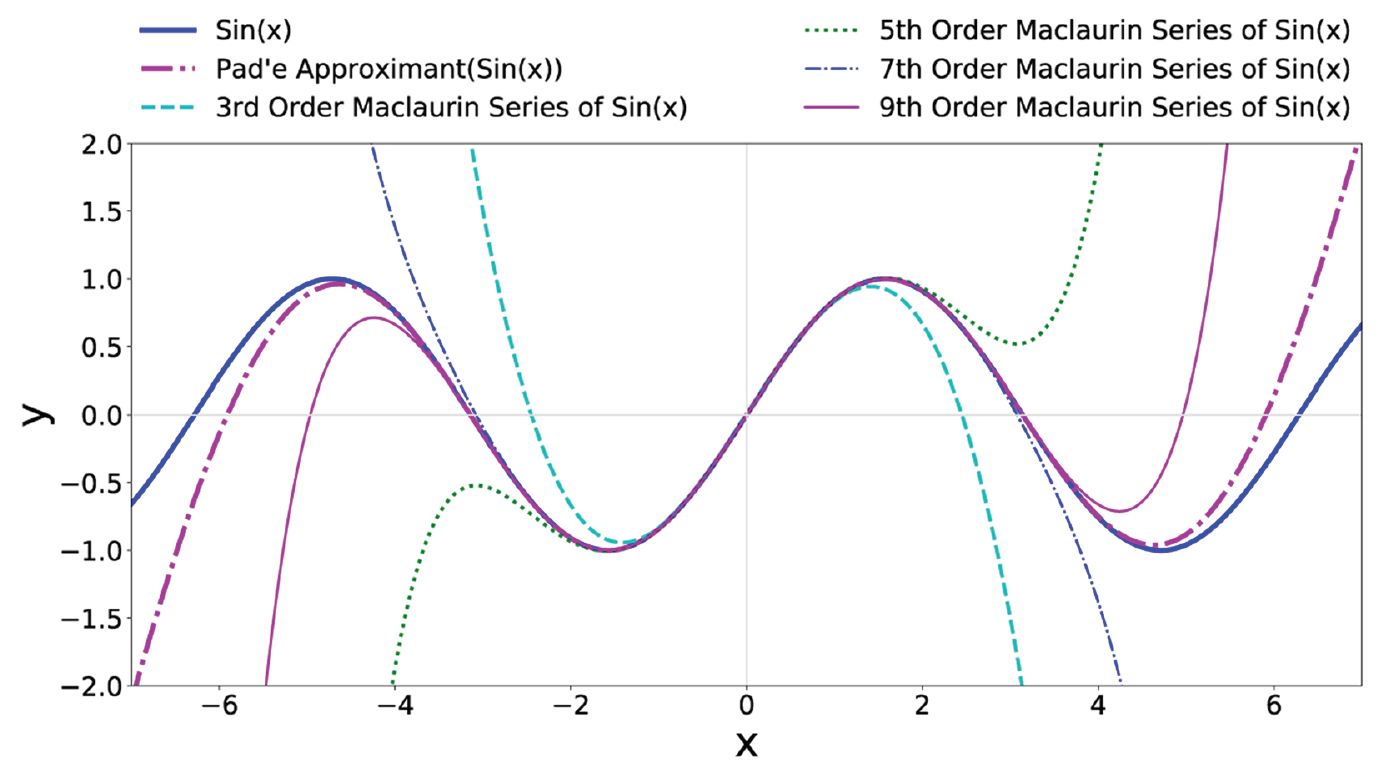}
    \caption{Approximation of $Sin(x)$ using the Pad\'e approximant of
     \eqref{eqn.pade-approximant-of-sin-function}
    and Maclaurin series expansion of $Sin(x)$ for orders=$\{3,5,7,9\}$. The values in $x$-axis are shown in $[-7,7]$ and $y$-axis are shown within the range of $[-2.0, 2.0]$.}
    \label{fig:sin_x_approx}
\end{figure}

We also know that every meromorphic function has best approximating rational functions known as the \textit{Pad\'e approximants}~\cite{Scholarpedia:Pade}. 
This is very interesting since it is known that, numerically, Pad\'e approximants are generally a better approximation to a function than truncating its Taylor series at the same order, and it may still be of use where the Taylor series does not converge. For example, the Pad\'e approximant of $\sin(x)$ at order [5/6] is given by:
\begin{equation}
    \sin(x)\approx {\frac {\frac{12671}{4363920}x^{5}-\frac{2363}{18183}x^{3}+x}
    {\frac{121}{16662240}x^{6}+\frac{601}{872784}x^{4}+\frac{445}{12122}x^{2}+1}},
    \label{eqn.pade-approximant-of-sin-function}
\end{equation}
which provides a close fit in the interval (-5,5) of the real numbers (see Fig.~\ref{fig:sin_x_approx}). 
A Maclaurin series expansion (i.e. a Taylor series centered at zero) would
need to be a polynomial of at least order 13 to approximate $\sin(x)$ to be at least competitive with expression given in~\eqref{eqn.pade-approximant-of-sin-function},
a polynomial of order 5 in the numerator and one of order 6 in the denominator.

For a problem involving symbolic regression in a multi-dimensional domain $D$, a possible representation could be one based on \textit{multivariate Pad\'e approximants}~\cite{DBLP:conf/issac/Chaffy86,DBLP:journals/moc/ZhouCT09,DBLP:journals/amc/AkalL18}. These representations may be useful since they may allow us to just use a basic mathematical form (ratio of multivariate polynomials). However, searching for a best fitting Pad\'e approximant would require us to guess a priori the degree of two polynomials (in the numerator and denominator); we offer a novel approach to circumvent this problem. Most remarkably, this new method for machine learning and artificial intelligence has roots in eighteen-century mathematics as we will show in the following section.

\subsubsection{Representation via analytic continued fractions}
We shall employ a representation based on regular C-fractions~\cite{DBLP:journals/toms/BackeljauwC09}, which are the corresponding continued fractions of a Pad\'e approximant~\cite{cfraction}.
To motivate this representation, we first note that the Pad\'e approximant for $\sin(x)$ given by~\eqref{eqn.pade-approximant-of-sin-function} can be written as follows:
\begin{equation}
\sin(x)\approx  x - \cfrac{\frac{1}{6}x^3}{1+\cfrac{\frac{1}{20}x^2}{1-\cfrac{\frac{11}{420}x^2}{1+\cfrac{\frac{25}{2772}x^2}{1-\cfrac{\frac{11}{900}x^2}{1+\cfrac{\frac{1331}{82650}x^2}{1}}}}}}
\end{equation}

Some readers may remember a result by Euler~\cite{Euler_1748} who first
derived an equation that connects a finite sum of products with a finite continued fraction,  as in Eq.~\eqref{Euler-finite-sum-of-products-formula}~\cite{Euler_1748}.
Let $SP=a_{0}+a_{0}a_{1}+a_{0}a_{1}a_{2}+\cdots +a_{0}a_{1}a_{2}\cdots a_{n}$, then:
\begin{equation}
SP={\cfrac  {a_{0}}{1-{\cfrac  {a_{1}}{1+a_{1}-{\cfrac  {a_{2}}{1+a_{2}-{\cfrac  {\ddots }{\ddots {\cfrac  {a_{{n-1}}}{1+a_{{n-1}}-{\cfrac  {a_{n}}{1+a_{n}}}}}}}}}}}}}\,
\label{Euler-finite-sum-of-products-formula}
\end{equation}
Since the result can be proved by induction on $n$, this means that we can apply this result in the limit. 

This fact takes directly to our proposed representation (e.g. the power of this representation for special functions can be found in \cite{DBLP:journals/toms/BackeljauwC09}). If, for instance, the left-hand side of the equation is a convergent and infinite series, then the right-hand side represents a \textit{convergent and infinite continued fraction}. As an example, we consider the function $\tanh(x_1 + 5 x_2)=\sinh(x_1 + 5 x_2)/\cosh(x_1 + 5 x_2)$.
Unless $\sinh(x_1 + 5 x_2)$ and $\cosh(x_1 + 5 x_2)$ are given as ``building blocks'', it would be very difficult for a system to evolve the ratio. However, Gauss proved in 1812 (see \cite{olds:1963:continued} p. 138, \cite{wall:1948:analytic} p. 349) that $\tanh(x_1 + 5 x_2)$ can be well written as:
\begin{equation}
\tanh(x_1 + 5 x_2) = 
\alpha_0 \, + \, \cfrac{f_0(\mathbf{x})}
{\alpha_1 + \cfrac{f_1(\mathbf{x})}
	{\alpha_2 + \cfrac{f_2(\mathbf{x})}
 		{\alpha_3 + \ddots}}}
\end{equation}
with $\alpha_0=1$, $f_0(\mathbf{x})=x_1 + 5 x_2$, $f_i(\mathbf{x})=(x_1 + 5 x_2)^2$, for any natural number $i > 0$, and with the set $\{\alpha_i\}$ with $i>0$ being the odd natural numbers in increasing values 
(i.e. $\alpha_1=1, \alpha_2=3, \alpha_3=5, \alpha_4=7, \alpha_5=9, \alpha_6=11, \dots$). Numerically, just a few levels would give a good approximation to the original function. 

With these examples in mind, we introduce a representation of mathematical expression which we will use for symbolic regression.
For a multivariate function $f : \mathbb{R}^n \to \mathbb{R}$ we can then write:

\begin{equation}
f(\mathbf{x}) = 
g_0(\mathbf{x})\, + \, \cfrac{h_0(\mathbf{x})}{g_1(\mathbf{x}) + \cfrac{h_1(\mathbf{x})}{g_2(\mathbf{x}) + \cfrac{h_2(\mathbf{x})}{g_3(\mathbf{x}) + \ddots}}}
\label{general-equation-for-CF}
\end{equation}
where $g_i(\mathbf{x}) \in \mathbb{R}$ for all integer $i\geq 0$.

For each function $f_i : \mathbb{R}^n \to \mathbb{R}$ is associated with a vector $\mathbf{a_i} \in \mathbb{R}^n$ and a constant $\alpha_i \in \mathbb{R}$:
\begin{equation}
g_i(\mathbf{x}) = 
\mathbf{a_i}^\mathrm{T} \mathbf{x} + \alpha_i
\label{eqn:for-g}
\end{equation}
Also, for each function $h_i : \mathbb{R}^n \to \mathbb{R}$ is associated with a vector $\mathbf{b_i} \in \mathbb{R}^n$ and a constant $\beta_i \in \mathbb{R}$:
\begin{equation}
h_i(\mathbf{x}) = 
\mathbf{b_i}^\mathrm{T} \mathbf{x} + \beta_i 
\label{eqn:for-h}
\end{equation}

From \eqref{general-equation-for-CF}, \eqref{eqn:for-g} and \eqref{eqn:for-h}, 
we can write (using Carl Friedrich Gauss' mathematical notation for continued fractions~\cite{DBLP:journals/toms/BackeljauwC09}):
\begin{equation}
    f(\mathbf{x}) = \mathbf{a_0}^\mathrm{T} \mathbf{x} + \alpha_0 + \K_{i=1}^{\infty} \left( \frac{\mathbf{b_i}^\mathrm{T} \mathbf{x} + \beta_i }{\mathbf{a_i}^\mathrm{T} \mathbf{x} + \alpha_i}\right)
    \label{eqn:Big-K-form-of-our-models}
\end{equation}
We then note that the 
Pad\'e approximant for $\sin(x)$ initially given by \eqref{eqn.pade-approximant-of-sin-function} can be written as: 
\begin{equation}
    \sin(x)\approx 
    x + \K_{k=1}^{6} \frac{
        \left\{ 
            \begin{array}{ll}
                {-x^3}/{6} & \mbox{$k=1$}\\
                {x^2}/{20} & \mbox{$k=2$}\\
                {-11x^2}/{420} & \mbox{$k=3$}\\
                {25x^2}/{2772} & \mbox{$k=4$}\\
                {-11x^2}/{900} & \mbox{$k=5$}\\
                {1331x^2}/{82650} & \mbox{$k=6$}
            \end{array} 
        \right.}{1}
        \label{eqn:Pade-for-sin-x-in-Big-K-format}
\end{equation}

This shows that the representation naturally contains the Pad\'e approximant as a finite sum in terms of the variables $x$ and $x^2$. 
Also, our representation allows a good truncated approximation to $\tanh()$ (e.g. by using Lambert’s continued fractions) even if the functions $\sinh()$ and $\cosh()$ have not been provided as ``building blocks''. Since only the four fundamental arithmetic operations are used, the continued fraction representation is able to approximate a function to increasing precision with more ``depth'' and in the limit it will converge to the target function of interest. We will formally define the notion of depth in Section~\ref{sec:depth}.

Analytic continued fractions are thus giving us a powerful representation; not only we can represent elementary functions (like $\sin(), \log(), \cos(), \tan^{-1}()$, etc.), but other special functions (like the Error function $\operatorname{erfc}()$, and the Incomplete Gamma Function, etc.) can also be represented in this way~\cite{Crandall:1994:PSC:156625}. We also note that there exists other 
possible representations (e.g. Stieltjes fractions, Thron fractions or Generalized T-Fractions, Thiele interpolating continued fractions, and Jacobi or J-Fractions)~(see~\cite{DBLP:journals/toms/BackeljauwC09} for their definitions). While exploring these alternative representations is an interesting research avenue with merits of its own, we concentrate our study in the one proposed here as a first benchmark on its performance for symbolic regression on a large dataset.

The use of analytic continued fractions would certainly power many other aspects of evolutionary computation due to its well-established mathematical properties and theoretical foundations. It has also the added advantage that it liberates from the need of sometimes ``guessing building blocks'' as continued fractions can represent an infinite number of functions that can be represented by an infinite but convergent series. Indeed, in addition to symbolic regression, we feel that analytic continued fractions will be a welcomed representation for the evolutionary computation community for many modelling approaches.

\subsubsection{Convergents and the depth of the representation}
\label{sec:depth}
While our representation is defined as an infinite sum, in practice, we expect that the sum should be truncated in a computational setting. Some solutions include: either we leave the truncation as another variable (since it directly relates to the ``complexity'' of the model that fits the data in question), or we leave it variable and somehow penalize those models that fit the data well but the truncation has a high value (e.g. for 
Eq.\eqref{eqn:Pade-for-sin-x-in-Big-K-format} that value is 6, and we will say that the ``depth'' is equal to 6).  Our formal definition of ``depth'' is associated with the convergent of a continued fraction. This links with known results from analytic continue fractions since the convergents of regular C-fractions are exactly the Pad\'e approximants~\cite{cfraction}, so the \textit{depth of $n$} for a model is defined to be the $n$th convergent of our representation (i.e. a model of depth of $n$ also contains a model of depth $n'$ for all $0 \leq n'< n$).
This said, if we decide to have a representation with a fixed ``depth'' of six (or the third convergent), this means that in Eq.~\eqref{general-equation-for-CF}, we have $h_6(\mathbf{x})=0$ but $g_5(\mathbf{x})$ is not constrained to be equal to zero.
Clearly, the selection of a higher order (and fixed) depth would guarantee a 
more precise fit, but may lead to overfitting on the training set and poor generalization in the test sets. 

Our representation is then quite general and, in principle, may be adopted for any arbitrarily defined maximum depth. For some problem domains, a multi-objective approach could be employed (e.g. following similar experiences in GP-based symbolic regression~\cite{DBLP:journals/tec/VladislavlevaSH09}), in which minimization of depth is one of the objectives since it directly contributes to model complexity.  We note that the use of model complexity as a second objective is common in GP and in other domains (e.g. to control so-called the undesired effect of ``bloating'' in the solutions obtained). Since solving the high-dimensional non-linear optimization problem we face, we need to employ a powerful and robust optimization solving technique, one that enforces both individual and global non-linear optimization.
Towards this end, we present our memetic approach in the following section.

%
%

\subsection{A memetic algorithm for model identification}
\label{sec:method}
In this section, we present a memetic algorithm~\cite{DBLP:series/sci/2012-379} to search for the best coefficients for~\eqref{eqn:Big-K-form-of-our-models}, and deliver a best model for our training data. We call this regression method using this memetic algorithm and continued fraction representation as `Continued Fraction Regression' (CFR). Our selection and design of this optimization technique in part stems from our familiarity with the technique~\cite{pablo} and the robustness shown in thousands of different applications, including many in the area of non-linear optimization.  We also aim here to show how it can be easily implemented from the combined use of non-linear optimization solvers and a few primitives that organize the search via interaction of a set of agents. 

The ``guiding function'' of a solution is its mean squared error (MSE) and the goal of the population is to find the model that minimizes the MSE. Clearly, others can be used as well. However, the comparison of different guiding functions is outside of the scope of this paper.

The advantage of the memetic algorithm is that it combines problem domain information (i.e. the use of CFR, based on established mathematical theory for function approximation), together with local optimizers which are robust and global search mechanisms to search for good subsets of variables to create models with. 

Unlike the GP's tree-based representation, we have a fixed representation. We do not search in a space of trees but in the high-dimensional space defined by all the coefficients in the formula of a CFR. In some sense, we are solving two optimization problems, one nested in the other one: 
(1) identify which are the variables that are needed in a model, and (2), given those variables what are the values of the coefficients in~\eqref{eqn:Big-K-form-of-our-models}. 

The decision of including or not a variable in turn decides if its associated coefficients should be optimized or not. This is handled by the recombination and mutation parts of the algorithm. 
A population of models is then maintained, and we refer as a ``generation'' the period of the MA evolutionary process in which we perform the operations of mutation, recombination and individual search optimization to find the best coefficients.
We employ a very simple recombination approach needed for variable selection, but our recombination could also be considered ``memetic''~\cite{10.1007/978-3-540-24855-2_57}.
We use a direct search method for individual model optimization and it is described in Sec.~\ref{individual-optimization-via-Nelder-Mead}. 
Coefficients are optimized by the use of a variant of a Nelder-Mead algorithm (which provides a kind of individual search mechanism). 

One key aspect we maintain from previous successful implementations of memetic algorithms~\cite{tsp} is the use a population structure, which is explained in 
Sec.~\ref{tree-hierarchical-structure}. 

We then start with the discussion of the individual model optimization and the population structure which may be the most uncommon feature of an MA for some of our readers. To improve the readability of the paper, we have shown the high-level view of the memetic framework for the symbolic regression in Algorithm~\ref{alg:cfr} and it summarizes the description of the algorithm presented in later sections.

%
\begin{algorithm}[!ht]
\DontPrintSemicolon
\SetAlgoLined
\setstretch{1}
\SetKwInOut{Input}{Input}\SetKwInOut{Output}{Output}
\Input{Num. of Vars $nVars$, Mutation Rate $\mu_{r}$, training data $D_{trn}$}
\Output{Best solution to fit the problem, $best$}
\BlankLine
 \tcc{Generate Initial Population by a  randomized  algorithm}
 $pop \leftarrow$ InitialPopulation($nVars, D_{trn}$)\;
 $best \leftarrow pocket(pop.root)$\;
 \BlankLine
 \For{$gen \in 1 \to numGen$}{
    \tcc{Mutate each current solution in the population}
    $pop_{\mu} \leftarrow$ Mutate($pop, \mu_{r}$)\;
    \tcc{Generate new Population by recombination mechanism}
    $pop_{r} \leftarrow$  RecombinePopulation($pop_{\mu}$)\;
    \BlankLine
    \tcc{Local Search Optimization of current solutions}
    \ForEach{$agent \in pop_r$}{
        LocalSearch($agent$)\;
    }
    \BlankLine
    \If{guiding function stagnates for consecutive 5 gens}{
        reset($pop_r.root$)\;
    }
    \tcc{Replace old population with evolved population}
    $pop \leftarrow pop_r$\;
    
    \BlankLine
    \tcc{Keep track of the best solution}
    \If{$guiding\_function(best) < guiding\_function(pocket(pop.root)) $}{
        $best \leftarrow pocket(pop.root)$\;
    }
 }
 return {$best$}\;
\caption{CFR Algorithm\label{alg:cfr}}
\end{algorithm}
\smallskip

\subsubsection{A memetic algorithm with a hierarchical population structure}
\label{tree-hierarchical-structure}

\paragraph{Tree-structured MAs} In fact, a tree-based population structure of agents was initially proposed in the early 90s~\cite{Moscato94blendingheuristics} and it has been subsequently used for the solution of many combinatorial optimization problems such as the asymmetric traveling salesman problem~\cite{Moscato94blendingheuristics,tsp}, hierarchical clustering and 
phylogenetics~\cite{DBLP:conf/ppsn/CottaM02,COTTA200375},
multistage capacitated lot-sizing problem~\cite{BERRETTA200467},  
gene expression linear ordering microarray data\cite{DBLP:conf/evoW/CottaMGFM03}, 
Lot Sizing and Scheduling\cite{DBLP:books/sp/chiong2012/ToledoAFM12}, 
total tardiness single machine scheduling~\cite{Mendes2002165},
Quadratic Assignment Problem~\cite{DBLP:conf/cec/HarrisBIM15}, 
Gate Matrix Layout problem~\cite{DBLP:conf/iberamia/MendesFMG02},
3D Protein Structure Prediction~\cite{DBLP:journals/cor/CorreaBKD18} to mention just a few very challenging applications. Other problem domains include the number partitioning problem~\cite{Berretta:2004:EPM:982409.982414}, and many problems in bioinformatics and visualization~\cite{10.1007/3-540-36605-9_3,10.1371/journal.pone.0014468}
dealt with by our group over many years (in which tree-based memetic algorithms were used to solve particular sub-problems of interest). The number of publications that have used a ternary-tree topology exceeds 40 (according to Google Scholar) and it has been employed during the research work of several PhD Theses in different countries being an established alternative to other topologies in use. In~\cite{tsp}, a comprehensive study of 40 different tree topologies was conducted with population sizes varying from 85 to 7 agents (each having two solutions being considered). Many of them included binary trees of different lengths. One measure of merit for these experiments was the ``gap'' to the known optimal solutions on the set of instances, and the CPU time employed. Interestingly, from this analysis, the ternary tree topology was shown to have the best quality trade-off between solution quality and time, and we have subsequently adopted this topology in several other applications.

\begin{figure}
	\centering
	\includegraphics{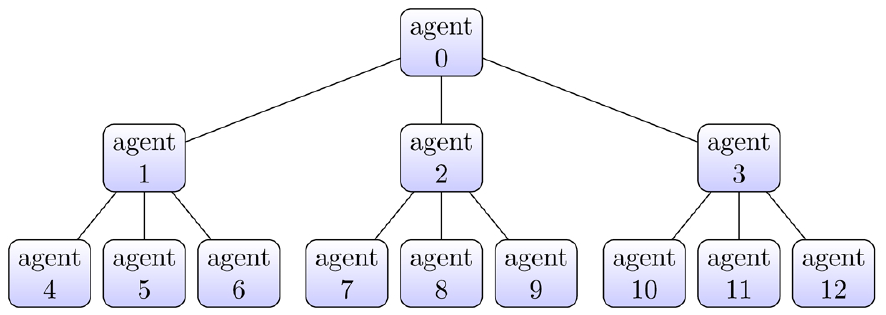} 
	\caption{The structure of a population consisting of 13 agents.}
	\label{fig:pop}
\end{figure}

\paragraph{Agents} The basic unit of population in our system is called an \textit{agent}. Each one has two possible different solutions (models), which are referred as the \textit{pocket} and the \textit{current}~\cite{tsp}\cite{DBLP:conf/iberamia/MendesFMG02}\cite{DBLP:conf/cec/HarrisBIM15}. Our population contains 13 agents and they are organized as a depth-3 ternary tree (Fig. \ref{fig:pop}). 
This approach has enabled in many cases good performances with a small population size of 13 agents (thus having a total of 26 solutions).

\paragraph{Initial Population} The initial population is decided by a randomized algorithm that works as follows. For each agent in the population we need to initialize two solutions. Each one is done the same and it is as follows. We select with equal probability the inclusion of a variable in a model, and when this is done we select at random with uniform probability a coefficient for each of the occurrences of a variable in a model (i.e. if we are working with depth equal to 8 it will be 17 coefficients to set to a value selected uniformly at random between -3 and 3 for each variable occurrence in the model). We also set in the same way any $\alpha_i$ and $\beta_i$ values in from \eqref{eqn:for-g} and \eqref{eqn:for-h}. 

Premature convergence in evolutionary algorithms needs to be addressed by introducing some techniques. Some of them are numerical, adaptive algorithms and heuristics that aim to control for loss of diversity. One that is non-numerical in nature is ``isolation by distance'', the introduction which are rules that prevent certain solutions to be recombined. One way of doing this is via the use of a ``population structure'', thus by restricting the recombination of solutions. 

Algorithm~\ref{alg:init-pop} shows the process of generating an initial population. Here, we generate 13 agents. We generate two solutions per agent (one `pocket' and one `current' solution), where each solution consists of a set of variables taken uniformly at random. Associated coefficients of the variables are also generated uniformly at random in the range of $-3$ to $3$. Then we evaluate the guiding function score of both solutions. We place the `pocket' and 'current' in such a manner so that it maintains the invariants of pocket solution having better guiding function value than the one of the current solution.

\paragraph{Invariants} Our algorithm maintains the following invariants in the population structure. 
First, within each agent, the \textit{pocket} solution always has a better value of the guiding function value than its \textit{current} solution. In addition, the pocket solution of each non-leaf node is always better or equal in quality to the three in the level immediately below to it. Maintaining these two invariants guarantees that there is a flow of good models upwards in the tree structure. 
A tree-based population structure avoids in part premature convergence~\cite{DBLP:journals/biosystems/MoscatoMB07}.

%
\smallskip
\begin{algorithm}
\DontPrintSemicolon
\SetAlgoLined
\setstretch{1}
\SetKwInOut{Input}{Input}\SetKwInOut{Output}{Output}
\Input{Number of Variables $nVars$, training data $D_{trn}$}
\Output{Randomly initialized Population $pop$}
\BlankLine
\tcc{Instantiate 13 Agents in the population}
$pop \leftarrow Agent(13)$\;
\ForEach{$agent \in pop$}{
    \tcc{Randomly generate two solutions per agent}
    $current(agent) \leftarrow$ ContinuedFraction($nVars$)\;
    $pocket(agent) \leftarrow$ ContinuedFraction($nVars$)\;
    \BlankLine
    \tcc{evaluate and save guiding function of both solutions}
    $guiding\_function(current(agent)) \leftarrow$ evalGF($current(agent), D_{trn}$)\;
    $guiding\_function(pocket(agent)) \leftarrow$ evalGF($pocket(agent), D_{trn}$)\;
    \BlankLine
    \tcc{within each agent, the pocket solution always holds better value of guiding function than its current solution}
    MaintainInvariant($agent$)\;
 }
 return {$pop$}\;
\caption{InitialPopulation\label{alg:init-pop}}
\end{algorithm}
\medskip

\subsubsection{Recombination and mutation to identify sets of variables for the models}
The evolutionary operators of mutation and recombination are responsible for the selection of the variables that will have nonzero coefficients and be included in a model.
For example, the following function
\begin{equation}
\label{equ:ex}
f(w, x, y, z) = 2.1 \: w + \cfrac{4.7 \: x + w +1.01}{x + \cfrac{1.3 + 5.7 \: y}{3.9 \: x}}
\end{equation}
only has the variable $z$ missing, so it uses three. 

The recombination operator takes two solutions and generates an offspring by combining the set of variables present in two parent individuals. At each step, one of the three possible recombination operators is chosen uniformly at random. We choose one recombination by considering the \textit{union} ($\cup$), \textit{intersection} ($\cap$), and 
\textit{symmetric difference} ($\triangle$) between the set of variables present in the models
to be recombined. 

The recombination operator then acts on each subpopulation, 
from top to bottom. Within each subpopulation, the operation is performed on each agent as follows (leader labelled as $\ell$, and supporters are labelled $s_1, s_2$ and $s_3$ from left to right):
\begin{enumerate}
    \item current($\ell$) $\gets$ Recombine(pocket($\ell$), current($s_1$))
    \item current($s_3$) $\gets$ Recombine(pocket($s_3$), current($\ell$))
    \item current($s_1$) $\gets$ Recombine(pocket($s_1$), current($s_2$))
    \item current($s_2$) $\gets$ Recombine(pocket($s_2$), current($s_3$))
\end{enumerate}


The \texttt{Recombine()} process of two agents to create an offspring is shown in Algorithm~\ref{alg:recombine}. Here, we select the variables in the offspring by applying the operator chosen uniformly at random on the variables of both parents. For each depth of the continued fraction, we compute the associated coefficient value for each variable. In the case of the new variable is being selected only from a single parent, where we copy the associated coefficient from that parent into the offspring solution. However, if this variable is already a part of the both parents ($a$ and $b$), we compute the new value of coefficient for corresponding variable as: $cof_a + rand(-1,4)*(cof_b - cof_a)/3$ (historically, this randomized approach has been previously used in early Scatter Search methods for non-linear optimization). We use the same formula to compute the value of constant ($\beta$) portion of the continued fraction. Then the guiding function is reevaluated. The new model then will contain a variable with nonzero coefficient only if that variable belongs to the set being preserved and at least one of the parents contains the same variable in the model. The individual search step is then the one in charge of optimizing the non-zero coefficients of the newly created model. 

%
\smallskip
\begin{algorithm}
\DontPrintSemicolon
\SetAlgoLined
\setstretch{1}
\SetKwInOut{Input}{Input}\SetKwInOut{Output}{Output}
\Input{Two agents take part in recombination $a, b$, choice of an operator uniformly at random form $\{\cup, \cap, \triangle \}$ as $OP$ }
\Output{Offspring as the Current solution of agent $a$}

\BlankLine
\tcc{Initialize parents ($p_1, p_2$) and offspring $ch$}
$p_1 \leftarrow pocket(a)$,
$p_2 \leftarrow current(b)$\;
$ch \leftarrow \phi$\;
\BlankLine

\tcc{Apply a recombination operator chosen uniformly at random on variables of two parents into offspring}
$ch.$vars()$ \leftarrow p_1$.vars() $OP$ $p_2.$vars()\;

\BlankLine
\tcc{Recombine the coefficients for each term ($h$) of the continued fraction}
\For{$i=0$ { to } $(2*depth)$}{
    $cof_a \leftarrow p_1.h_i$.coef(), 
    $vars_a \leftarrow p_1.h_i$.vars()\;
    
    $cof_b \leftarrow p_2.h_i$.coef(), 
    $vars_b \leftarrow p_2.h_i$.vars()\;
    \BlankLine
    \tcc{recombine coefficient values for variables}
    $cof_{\chi} \leftarrow \phi$\;
    \For{$vi=1$ to $nVars$}{
        \If{$ch$.featAt($vi$)$=true$}{
            \uIf{$vars_a[vi]=true$ and $vars_b[vi]=true$}{
                $cof_{\chi}[vi] \leftarrow cof_a[vi] + rand(-1,4)*(cof_b[vi] - cof_a[vi])/3$\;
            }\uElseIf{$vars_a[vi]=true$}
            {
                $cof_{\chi}[vi] \leftarrow cof_a[vi]$\;
            }\ElseIf{$vars_b[vi]=true$}{
             $cof_{\chi}[vi] \leftarrow cof_b[vi]$\;
            }
        }
    }
    \tcc{Update new coefficients of the term in offspring}
    $ch.h_i.coef() \leftarrow cof_{\chi}$

    \tcc{compute new value of constant ($\beta$) for term $h_i$ in the offspring solution $ch$ using $\beta$ of $p_1.h_i$ and $p_2.h_i$}
    $ch.h_i.\beta \leftarrow p_1.h_i.\beta + rand(-1,4)*(p_2.h_i.\beta - p_1.h_i.\beta)/3$\;
}
\BlankLine
\tcc{Update current solution and apply Local Search}
$current(a) \leftarrow ch$\;
$a \leftarrow localSearch(a)$\;
return $a$\;
\caption{Recombine\label{alg:recombine}}
\end{algorithm}
\medskip 

Like in other evolutionary computation algorithms, mutation is a random mechanism used to increase the diversity of the population. Here we have decided to incorporate two forms of mutation operators (`major mutation' and `soft mutation') in our implementation to toggle random variables. We choose a mutation operation depending on the guiding function value of current and pocket solutions. We do a `major mutation' on the current \textit{if the guiding function value is either within 120\% of the guiding function value of pocket ($current.guiding\_function < 1.2 * pocket.guiding\_function$) } \textit{or greater than twice of the pocket's guiding function value} ($current.guiding\_function > 2 * pocket.guiding\_function$). A `soft mutation' is taken place in other cases.

For the `major mutation' (detail of this mutation operation, \texttt{toggleVariables()}, is shown in Algorithm~\ref{alg:mut-maj}), we select a variable uniformly at random for toggle and modify associated coefficients in all depths of the continued fraction. If the randomly selected variable was already  incorporated (``switched on'') for the depth, we remove the variable. However, either we ``remember'' the existing coefficient value or assign `zero' to ``remove'' with probability 50\%. Alternatively, if the variable was not incorporated, we switch it on and replace the coefficient either by `zero' or by a uniformly random number in the range of -3 to 3 with the probability of 50\%.
In the case of `soft mutation' (shown in Algorithm~\ref{alg:mut-minor} as \texttt{modifyVariable()}), we select a depth and a variable, both uniformly at random. If the variable was already incorporated, toggle the variable selection and `remove' the coefficient by assigning zero as value. Otherwise, we `modify' the coefficient of the random variable by a uniformly generated random number in the range of -3 to 3 and toggle the variable selection.
Finally, the local search operation is executed on the mutated solution in order to optimize non-zero coefficients. 
We do not apply mutation on pocket solutions because we consider them as a ``collective memory'' of good models visited in the past. They influence the search process via recombination only.

%
\smallskip
\begin{algorithm}
\DontPrintSemicolon
\SetAlgoLined
\setstretch{1}

\SetKwInOut{Input}{Input}\SetKwInOut{Output}{Output}
\Input{A Continued Fraction Solution $cfrac$}
\Output{The modified solution $cfrac$}
\BlankLine
\tcc{select a variable index uniformly at random}
$N \leftarrow cfrac.nVars$\;
$vIdx \leftarrow$ randChoice$(N)$\;
\BlankLine
\tcc{for each depth of continued fraction, toggle the selection of variables of the term ($h$)}
\ForEach{$h \in cfrac$}{
    \tcc{Case 1: cfrac variable is turned ON: Turn OFF the variable, and either `Remove' or `Remember' the coefficient value at random}
    \eIf{$cfrac$.varAt$(vIdx) = true$}{
        $h$.varAt$(vIdx) \leftarrow false$\;
        $h$.coefAt$(vIdx) \leftarrow$ coinToss$(0, h.$coefAt$(vIdx) )$\;
    }{    
        \tcc{Case 2: term variable is turned OFF: Turn ON the variable, and either `Remove' or `Replace' the coefficient with a random value between -3 to 3 at random}
        \If{$h$.varAt$(vIdx) = false$}{
            $h$.varAt$(vIdx) \leftarrow true$\;
            $h$.coefAt$(vIdx) \leftarrow$ coinToss$(0$, rand(-3 3))\;
        }
    }
}
\tcc{Toggle the randomly selected variable}
$cfrac.$varAt$(vIdx) \leftarrow \lnot cfrac.$varAt$(vIdx)$\;
\caption{toggleVariables\label{alg:mut-maj}}
\end{algorithm}
\smallskip

%
\smallskip
\begin{algorithm}
\DontPrintSemicolon
\SetAlgoLined
\setstretch{1}
\SetKwInOut{Input}{Input}\SetKwInOut{Output}{Output}
\Input{A Continued Fraction Solution $cfrac$}
\Output{The modified solution $cfrac$}
\BlankLine
\tcc{Randomly select a variable which is turned ON}
$candVars \leftarrow \{\forall i : cfrac.$varAt$(i)=true\}$\;
$vIdx \leftarrow$ randChoice$(candVars)$\;
\BlankLine
\tcc{Randomly select a term ($h$) of continued fraction}
$h \leftarrow $ randChoice$(\{\forall term \in cfrac\})$\;

\BlankLine
\tcc{Modify the coefficient value}
\eIf{$h.$varAt$(vIdx)=true$}{
  $h.$coefAt$(vIdx) \leftarrow 0$\;  
}{
    $h.$coefAt$(vIdx) \leftarrow$ rand(-3 3)\; 
}
\tcc{Toggle the randomly selected variable}
$h.$varAt$(vIdx) \leftarrow \lnot h.$varAt$(vIdx)$\;

\caption{modifyVariable\label{alg:mut-minor}}
\end{algorithm}
\smallskip

\subsubsection{Individual model optimization via a direct search method}
\label{individual-optimization-via-Nelder-Mead}
 
A period of individual search operation is performed every generation on all current solutions. We remind again that each solution corresponds to a single model, this means that if a current model becomes better than its corresponding pocket model (in terms of the guiding function of the solution), then an individual search optimization step is also performed on the pocket solution/model before we swap it with the current solution/model. Individual search can then make a current model better than the pocket model (again, according to the guiding function), and in that case they switch positions within the agent that contains both of them. 

To do these optimizations, we used a modified version of Nelder-Mead algorithm recently proposed by Fajfar \etal~\cite{nelder-mead} to optimize the coefficients of a model. Nelder-Mead methods, also known as the \textit{downhill simplex algorithm}, is a derivative-free nonlinear optimization algorithm known for its simplicity and relatively good empirical performance~\cite{Singer:2009}.

To run the Nelder-Mead algorithm, the list of constants and coefficients in a model is mapped to a vector by the order in which they appear.
The initial simplex is generated by adding a unit step to each dimension of said vector. 
This heuristic of initializing the simplex is described in~\cite{Singer:2009}.
The individual search stops when the best and worst vertex of the simplex are within numerical tolerance, or when a maximum number of iterations is reached.
In our implementation, we set the numerical tolerance as $10^{\minus 3}$ and the maximum iterations of 250.

While our memetic algorithm does individual search optimization with a modified Nelder-Mead algorithm~\cite{nelder-mead}, the nature of the non-linear optimization problem indicates that more powerful solvers could later be used. We have chosen to start with a Nelder-Mead solver as a way of ensuring and promoting reproducibility, keeping the core memetic algorithm as simple as possible~\cite{HaoMoscato2019:cec} for this first in-depth test of performance of the new representation in real-world problems.

\subsubsection{Small batch learning}
\label{sec:small-batch}
To deal with datasets having more than 200 samples, during an execution of individual search, a small subset of the set of training samples (20\%) is selected uniformly at random (from the whole training set) and during the individual search optimization, the guiding function value is computed only using these samples. This selection is conducted every time an individual search optimization is required. For each model, 4 independent local searches are performed (using the variant of the Nelder-Mead algorithm discussed before); the result with the best guiding function value (on the entire dataset) is chosen. This allows to have some sort of sampling of the quality of the variables in independent trials of the individual search process. 

\subsubsection{Diversity Management}
\label{sec:diversity}
We have set up a time-limit on the best model currently in the population (i.e. the one represented by the pocket solution of the root agent of our tree hierarchy) to influence the search. Our criterion for relevance has been fairly strict: if no better model has been produced for five (5) straight generations, then the pocket of the root agent is removed and a new solution is created at random. This is a fairly strong requirement, but for a depth-3 ternary tree, a solution that has climbed to the top and has been ``the best seen'' for five generations had already the opportunity to influence the search procedure and we avoid being trapped into a local minimum of the search space and not exploring other combinations of variables.

\subsubsection{Model complexity management}
\label{sec:complexity}
In the area of symbolic regression, and in some GP implementations, the tendency of fitting better the data at the cost of producing more complex models is called ``bloating''. 
This is an undesirable characteristic as one of the objectives of symbolic regression is to have easy to interpret, small and useful models and with better generalization~\cite{DBLP:conf/seal/Dick14}. 

We decided to set up an adaptive control for the complexity of our feasible solutions by penalizing the number of variables being used by a model. At initialization of the set of models, a whitelist is generated uniformly at random and independently done for each individual.
Each input variable then has a probability $p=1/3$ to be present in the whitelist. 
Only variables on the whitelist may be assigned with a value so that no bloated solution appear in the initial population.
To bias the search towards solutions of lesser complexity, we penalize the complexity.

As mentioned above, the \textit{Mean Squared Error} (MSE) metric is used to quantify the goodness of solution. It is computed for the average of the squared error of prediction (${y}'$) with the measured/observed (${y}$) value of the target for all $n$ samples $\{(\mathbf{x^{(i)}}, y^{(i)})\}$ in the dataset ($S$) according to:
\begin{equation}
    MSE = \frac{1}{n}\sum_{i=1}^{n}{(y^{(i)} - {y'}^{(i)})}^2.
    \label{eq:mse}
\end{equation}

Instead of comparing the quality of the solutions by the goodness of fit alone (measured by MSE), we aim at minimizing an ``adjusted MSE'' given by:
\begin{equation}
\text{adjusted MSE} = \text{MSE} \times (1 + \Delta \times \text{\# of variables used})
\label{eq:guidingfunction}
\end{equation}
where $\Delta > 0$ is the scale of the penalty.
Different values of $\Delta$ help to achieve different balances between goodness of fit and complexity depending on the workload.

Table~\ref{tab:param} shows the value of the parameters used for our experiments.
\begin{table}
 \centering

 \caption{Parameter value of the CFR-based memetic algorithm for regression.\label{tab:param} }
 \footnotesize{
\begin{tabular}{lr}
\toprule
Parameter & Value \\
\midrule
Delta & 0.10\\
Fraction's Depth & 4\\
Reset root of population after stuck for generations & 5\\
Number of Generations & 200\\
Mutation Rate & 0.10\\
Number of Nelder-Mead Instances & 4\\
Number of Iterations in Nelder-Mead & 250\\
Nelder-Mead terminates if stagnates for consecutive iterations of & 10\\
Percentage of Samples used to evaluate a model in local search & 20\% \\
\bottomrule
\end{tabular}
}
\end{table}

\subsection{Learning the Gamma Function\label{sec:learning-the-gamma-function}}
\begin{figure}
\centering
\subfigure[$Depth=2$]{\includegraphics[width=0.6\columnwidth]{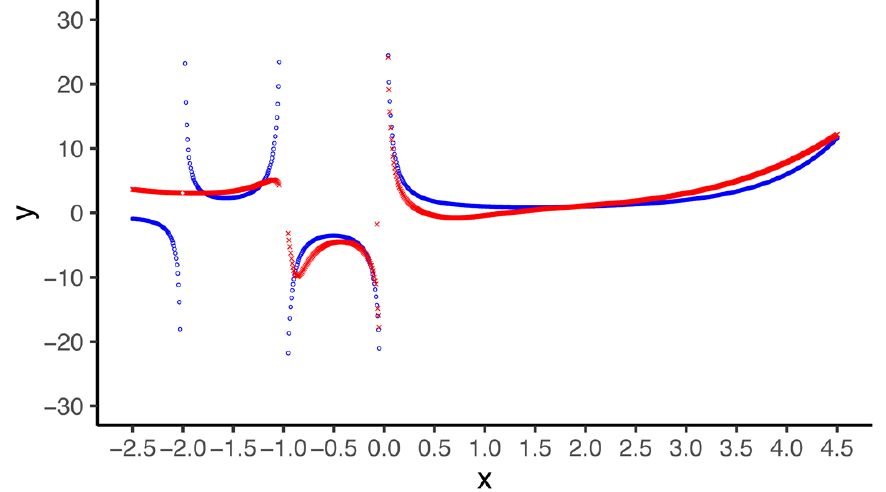}%
\label{sfig_depth_2}}
\subfigure[$Depth=4$]{\includegraphics[width=0.6\columnwidth]{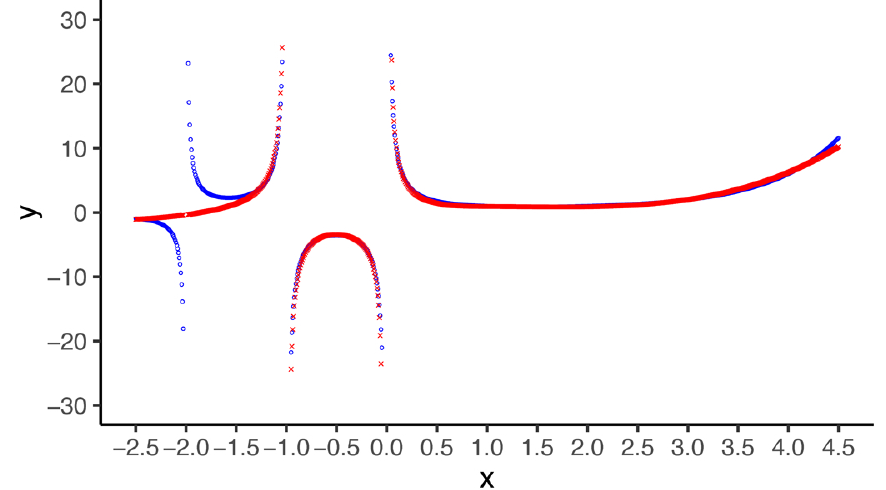}%
\label{sfig_depth_4}}
\subfigure[$Depth=6$]{\includegraphics[width=0.6\columnwidth]{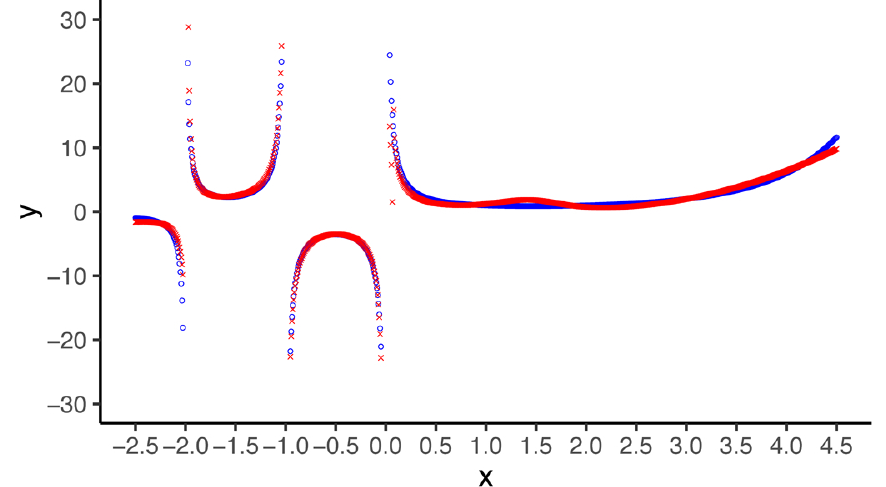}%
\label{sfig_depth_6}}
\caption{Results of our memetic algorithm (red +) learning the
$\Gamma(x)$ function with $x \in [-2.683, 4.5]$ (blue o) producing models restricted to a maximum value of Depth:  (a) $Depth=2$, (b) $Depth=4$ and (c) $Depth=6$ in CFR. The values in $Y$-axis are shown in the range of $\{-30,\cdots,30\}$. However, we have two points at $x=-1.978$ and $x=0.072$ out of this range ($y=62.00324026$ and $y=139.476602$, respectively) for $Depth=6$ (in Sub-Fig. (c)) that have been omitted for clarity. Understandably, with a deeper model we can approximate the function better, although we have observed some runs for which $Depth=4$ obtaining results remarkably similar to those of $Depth=6$ in this figure. This indicates that global optimization methods will have a remarkable role in using small depth analytic continued fraction representations for symbolic regression.}
\label{fig_example_depth}
\end{figure}

The Gamma function $\Gamma(x)$ is one of a number of analytic extensions of the factorial function to real and complex numbers. Many common integral calculations in applied mathematics can be expressed in terms of the Gamma function.
For complex numbers with a positive real part, the Gamma function is defined via the convergent improper integral:
\begin{equation}
\Gamma (z)=\int _{0}^{\infty }x^{z-1}e^{-x}\,dx. 
\end{equation}
However, the only complex numbers for which the $\Gamma(z)$ is not defined are the non-positive integers for which it has simple poles. 

The Gamma function occurs in many areas of statistics and data analysis and it frequently appears in the study of natural phenomena where there exists a process in which an decay in time is present and this decay follows a law of the form $f(t) e^{-g(t)}$ where $t$ represents time~\cite{miller2015.chapter15.Gamma}.  For instance, when $f(t)$ is a power function and $g(t)$ is a linear function of time, respectively, thanks to a change of variables we can write:
\begin{equation}
\int _{0}^{\infty }t^{b}e^{-at}\,dt={\frac {\Gamma (b+1)}{a^{b+1}}}.
\end{equation}

We have chosen to show the performance of the technique on a testbed problem before presenting the computational results in a large benchmarking dataset. This illustrates the role of the depth of the continued fraction representation in increasing the approximation to the true values. This said, we propose that learning the values of $\Gamma(x)$ on an interval on the reals around the zero is an interesting challenge for symbolic regression methods and could be now added to the list of benchmarking functions of one variable. It is also an interesting test function since for non-positive integers the target function values alternate signs between its poles, while for positive integers there are no poles but its values increase faster than an exponential function (since for any positive integer $n$, $\Gamma(n)=(n-1)!$). 

As an example of how we can approximate the Gamma Function in the interval in which the function alternates signs between some poles, we have chosen the interval $[-2.683,4.5]$ (see Fig~\ref{fig_example_depth}). We have generated a set $S=\{(x,y)\}$ of 873 samples using a uniform separation step. To simulate a situation similar to the one we will observe in the experimental setting that was used later in the paper (in which we compare our results with other methods having a high-dimensional input space), we have created an input space consists of the variable $x$ and all of its powers up to degree 6. This then creates an instance of symbolic regression of 7 dimensions since now we have $\mathbf(x)=[1, x, x^2, \dots x^6]$. We have run our memetic algorithm for several depths of the representation, i.e. $2, 4$ and $6$. Fig.~\ref{fig_example_depth} shows the learning outcome of $\Gamma(x)$ for different depths.

We note that the GP based software Eureqa version 1.24.0~(\cite{Schmidt2009,Eureqa:Software}) was able to very quickly find 
a solution of the form:
\begin{equation}
\Gamma(x) \approx a x^3 + \frac{b}{x + x^2},
\end{equation}
with $a \approx 0.096350$ 
and $b \approx 1.00808$
However, further computation using nearly one million generations (approximately 40 minutes) was required to finally deliver a better solution:

\begin{equation}
\Gamma(x) \approx c_1 + c_2 x + \frac{c_3}{x} + c_4 x^6 + \frac{c_5}{c_6 x + x^3 + c_7 x^2},
\end{equation}

\noindent
with 
$c_1 \approx 0.201148$, 
$c_2 \approx 0.216747$,
$c_3 \approx 0.492681$,
$c_4 \approx 0.001204$,
$c_5 \approx 1.004562$,
$c_6 \approx 2.000514$,
$c_7 \approx 3.000165$.

In terms of time, our method required only 
87.5 seconds, a fraction of the time employed by Eureqa on the same laptop, to obtain the solution shown in Fig.~\ref{sfig_depth_6}. Since Eureqa cannot be used for large-scale batch experimentation in our clusters (being an interactive package running on our workstations), we resort to a battery of tests recently proposed for the study of the performance of genetic programming and other machine learning methods in systematic manner. The next section presents the benchmarking tests and results.

\subsection{Experimental settings and Datasets\label{sec:exp-ds}}
To compare the performance of our proposed algorithms with other state-of-the-art algorithms we have used a standard \textit{benchmark} dataset~\cite{PMLB:Olson2017}. In addition to MSE score, we will also compute the normalized MSE score (NMSE) that provides information about the deviation (loss) of the performances and which allows comparisons between different databases. The NMSE score is calculated by normalizing the MSE score with respect to the variance ($\Var$) of the measured/observed value ($y$)~\cite{Quino:eval-pred:2005} according to:
$\Var(y) = \frac{1}{n-1}\sum_{i=1}^{n}{ |y_i - \overline{y}|^2}$. The NMSE score is given by:
\begin{equation}
    NMSE = \frac{1}{n}\sum_{i=1}^{n}{ \frac{ \left(y_{i} - {y}^{'}_{i}\right)^2 }{ \Var(y) } }.
    \label{eq:nmse}
\end{equation}

\subsubsection{The 94 Datasets from the Penn Machine Learning Benchmarks\label{sec:datasets}}

Oslon~\etal~\cite{PMLB:Olson2017} compiled a collection of datasets to evaluate machine learning problems, known as \textit{Penn Machine Learning Benchmarks} (PMLB)\footnote{\url{https://github.com/EpistasisLab/penn-ml-benchmarks}}. Later on, Orzechowski~\etal~\cite{PMLB-Regression:GECCO2018} selected a subset of 94 datasets from PMLB which are suitable for regression problems. We have employed these 94 datasets~\cite{PMLB-Regression:GECCO2018} for the evaluation of our memetic algorithm.

Since each of those 94 datasets only has a single file of samples associated to the regression problem, we have selected uniformly at random 75\% of the entries as a training set and the remaining 25\% as testing  for the algorithm to measure the performance (i.e. an independent test set). To make the results more significant and free from bias, that split of train-test has been taken uniformly at random for each of the 100 independent executions of the memetic algorithm.

\subsubsection{Computational Environment\label{sec:exe-env}}
We have compiled the algorithm with \texttt{gcc} compiler version \texttt{4.8.5} in 64 bit Red Hat Enterprise Linux Server 7.5 (Maipo). We have executed the algorithm on The University of Newcastle's High Performance Computing (HPC) grid\footnote{\url{https://www.newcastle.edu.au/research-and-innovation/resources/research-computing-services/advanced-computing}}; we split the execution of the CFR on 94 datasets into 10 batches. For each batch, we assigned 2 CPU cores, 8GB RAM, and a total of 100 hours of CPU wall time to execute 100 independent runs of the algorithm.

%
%
\section{Results\label{sec:result}}

\subsection{Results on Penn Machine Learning Benchmarks Datasets}

We measured the MSE and NMSE scores obtained on the training set for each of the datasets. The model found during the training phase was also evaluated on the testing data. We tabulated the median scores of MSE and NMSE for both in training (\texttt{t}) and testing (\texttt{T}) scores of the algorithm for the 100 independent runs. Table~\ref{tab:penn-ml-100run} summarizes the scores for all benchmark datasets. We have sorted the results from the smallest to the largest of the median NMSE scores achieved in testing.

\footnotesize{

\begin{longtable}{
    >{\raggedright \arraybackslash}p{1.75cm}
    >{\raggedleft \arraybackslash}p{1cm}
    >{\raggedleft \arraybackslash}p{1cm}
    >{\raggedleft \arraybackslash}p{1cm}
    >{\raggedleft \arraybackslash}p{1cm} 
    >{\raggedright\arraybackslash}p{1.75cm}
    >{\raggedleft \arraybackslash}p{1cm}
    >{\raggedleft \arraybackslash}p{1cm}
    >{\raggedleft \arraybackslash}p{1cm}
    >{\raggedleft \arraybackslash}p{1cm}
    }

\caption{Median value of the proposed approach (Computed Over 100 runs) for the MSE and NMSE scores on Training (t.) and Testing (T.) splits on the 94 Benchmark Datasets.\label{tab:penn-ml-100run}}
\\

\hline\noalign{\smallskip}
{Dataset} & {$t_{mse}$} & {$T_{mse}$} & {$t_{nmse}$} & {$T_{nmse}$} & {Dataset} & {$t_{mse}$} & {$T_{mse}$} & {$t_{nmse}$} & {$T_{nmse}$}\\
\noalign{\smallskip}\hline\noalign{\smallskip}

\endhead
\noalign{\smallskip}\hline\noalign{\smallskip}
\multicolumn{10}{r}{\footnotesize Continue on the next page}
\endfoot
\endlastfoot

rabe\_266 & 8.381 & 9.641 & 0.003 & 0.004 & sl\_ex1714 & 1.11e6 & 1.86e6 & 0.145 & 0.299 \\
a.elec2000 & 4.09e6 & 1.42e7 & 0.001 & 0.005 & fri\_c3\_500\_10 & 0.238 & 0.278 & 0.244 & 0.307 \\
a.neavote & 0.725 & 0.903 & 0.046 & 0.059 & fri\_c2\_500\_50 & 0.270 & 0.310 & 0.284 & 0.307 \\
cpu & 6.37e2 & 1.33e3 & 0.028 & 0.077 & fri\_c0\_250\_5 & 0.249 & 0.308 & 0.246 & 0.311 \\
fri\_c2\_1000\_5 & 0.069 & 0.078 & 0.069 & 0.080 & fri\_c0\_1000\_50 & 0.275 & 0.313 & 0.279 & 0.314 \\
fri\_c1\_1000\_5 & 0.079 & 0.088 & 0.079 & 0.088 & fri\_c2\_250\_25 & 0.249 & 0.326 & 0.252 & 0.317 \\
fri\_c3\_1000\_5 & 0.081 & 0.094 & 0.080 & 0.096 & fri\_c0\_500\_10 & 0.285 & 0.324 & 0.281 & 0.324 \\
v.galaxy & 8.00e2 & 8.86e2 & 0.090 & 0.099 & fri\_c0\_500\_25 & 0.262 & 0.307 & 0.262 & 0.325 \\
fri\_c2\_500\_5 & 0.083 & 0.105 & 0.083 & 0.105 & fri\_c0\_500\_50 & 0.285 & 0.349 & 0.285 & 0.346 \\
fri\_c4\_1000\_10 & 0.103 & 0.115 & 0.106 & 0.112 & fri\_c1\_500\_50 & 0.333 & 0.355 & 0.342 & 0.347 \\
fri\_c1\_500\_5 & 0.095 & 0.116 & 0.096 & 0.117 & fri\_c1\_1000\_50 & 0.330 & 0.350 & 0.333 & 0.349 \\
a.apnea2 & 1.17e6 & 1.17e6 & 0.114 & 0.129 & sl\_case1202 & 2.17e3 & 2.80e3 & 0.262 & 0.355 \\
fri\_c1\_1000\_10 & 0.129 & 0.135 & 0.130 & 0.134 & fri\_c0\_250\_10 & 0.298 & 0.376 & 0.301 & 0.373 \\
a.apnea1 & 1.16e6 & 1.21e6 & 0.109 & 0.138 & fri\_c0\_100\_5 & 0.240 & 0.377 & 0.243 & 0.388 \\
chatfield\_4 & 2.47e2 & 2.96e2 & 0.123 & 0.141 & fri\_c3\_250\_10 & 0.343 & 0.413 & 0.339 & 0.420 \\
fri\_c3\_1000\_25 & 0.128 & 0.141 & 0.129 & 0.150 & fri\_c4\_250\_25 & 0.333 & 0.394 & 0.321 & 0.421 \\
ESL & 0.265 & 0.309 & 0.133 & 0.150 & fri\_c0\_250\_25 & 0.293 & 0.411 & 0.295 & 0.422 \\
fri\_c2\_1000\_10 & 0.136 & 0.157 & 0.138 & 0.151 & sl\_ex1605 & 95.070 & 103.057 & 0.411 & 0.423 \\
fri\_c1\_500\_10 & 0.138 & 0.153 & 0.140 & 0.153 & fri\_c0\_250\_50 & 0.345 & 0.431 & 0.340 & 0.423 \\
fri\_c2\_1000\_25 & 0.154 & 0.156 & 0.154 & 0.160 & fri\_c3\_250\_25 & 0.323 & 0.407 & 0.322 & 0.426 \\
fri\_c3\_500\_5 & 0.128 & 0.159 & 0.131 & 0.160 & rmftsa\_ladata & 3.135 & 3.208 & 0.384 & 0.433 \\
fri\_c1\_1000\_25 & 0.155 & 0.175 & 0.156 & 0.167 & a.vehicle & 2.17e4 & 3.73e4 & 0.279 & 0.440 \\
fri\_c4\_1000\_25 & 0.145 & 0.176 & 0.145 & 0.179 & fri\_c1\_250\_10 & 0.379 & 0.458 & 0.384 & 0.449 \\
fri\_c3\_1000\_10 & 0.156 & 0.168 & 0.154 & 0.180 & fri\_c3\_1000\_50 & 0.383 & 0.449 & 0.387 & 0.453 \\
FacultySalry & 1.751 & 3.127 & 0.083 & 0.180 & fri\_c2\_100\_5 & 0.328 & 0.500 & 0.326 & 0.459 \\
machine\_cpu & 2.24e3 & 3.74e3 & 0.089 & 0.190 & fri\_c1\_100\_10 & 0.365 & 0.472 & 0.369 & 0.462 \\
fri\_c1\_250\_5 & 0.128 & 0.187 & 0.126 & 0.190 & LEV & 0.409 & 0.418 & 0.444 & 0.465 \\
fri\_c2\_250\_5 & 0.147 & 0.187 & 0.145 & 0.190 & vineyard & 4.734 & 7.993 & 0.242 & 0.475 \\
cloud & 0.095 & 0.165 & 0.083 & 0.195 & fri\_c4\_1000\_50 & 0.423 & 0.445 & 0.417 & 0.476 \\
fri\_c0\_1000\_5 & 0.182 & 0.197 & 0.182 & 0.200 & fri\_c2\_100\_10 & 0.382 & 0.537 & 0.399 & 0.482 \\
fri\_c2\_500\_10 & 0.178 & 0.217 & 0.178 & 0.211 & fri\_c0\_100\_10 & 0.277 & 0.504 & 0.279 & 0.487 \\
fri\_c4\_500\_25 & 0.164 & 0.212 & 0.163 & 0.218 & fri\_c4\_500\_10 & 0.436 & 0.484 & 0.407 & 0.488 \\
auto\_price & 6.54e6 & 7.62e6 & 0.187 & 0.219 & fri\_c4\_1000\_100 & 0.526 & 0.541 & 0.520 & 0.546 \\
elusage & 86.700 & 1.19e2 & 0.147 & 0.219 & fri\_c1\_250\_50 & 0.457 & 0.532 & 0.452 & 0.551 \\
autoPrice & 6.35e6 & 7.56e6 & 0.184 & 0.220 & no2 & 0.303 & 0.320 & 0.531 & 0.563 \\
fri\_c3\_500\_25 & 0.177 & 0.242 & 0.177 & 0.225 & fri\_c3\_500\_50 & 0.582 & 0.605 & 0.582 & 0.595 \\
USCrime & 2.02e2 & 3.72e2 & 0.127 & 0.227 & fri\_c1\_100\_5 & 0.397 & 0.579 & 0.395 & 0.598 \\
fri\_c2\_1000\_50 & 0.214 & 0.230 & 0.214 & 0.230 & pollution & 1.29e3 & 2.40e3 & 0.319 & 0.623 \\
chs\_geyser1 & 36.539 & 38.619 & 0.223 & 0.237 & ERA & 2.553 & 2.665 & 0.653 & 0.671 \\
fri\_c2\_500\_25 & 0.219 & 0.228 & 0.217 & 0.242 & sl\_case2002 & 55.722 & 59.521 & 0.591 & 0.674 \\
vinnie & 2.375 & 2.371 & 0.253 & 0.242 & fri\_c4\_500\_50 & 0.625 & 0.664 & 0.623 & 0.680 \\
fri\_c2\_250\_10 & 0.204 & 0.239 & 0.210 & 0.248 & v.env. & 7.380 & 8.412 & 0.574 & 0.687 \\
fri\_c3\_250\_5 & 0.202 & 0.274 & 0.196 & 0.252 & fri\_c4\_250\_10 & 0.594 & 0.702 & 0.587 & 0.697 \\
fri\_c0\_500\_5 & 0.210 & 0.258 & 0.206 & 0.258 & fri\_c3\_100\_5 & 0.469 & 0.679 & 0.463 & 0.698 \\
fri\_c0\_1000\_10 & 0.254 & 0.260 & 0.253 & 0.262 & fri\_c0\_100\_25 & 0.450 & 0.778 & 0.461 & 0.745 \\
fri\_c0\_1000\_25 & 0.222 & 0.258 & 0.223 & 0.266 & SWD & 0.484 & 0.488 & 0.736 & 0.749 \\
fri\_c1\_500\_25 & 0.204 & 0.278 & 0.203 & 0.275 & pm10 & 0.681 & 0.693 & 0.870 & 0.885\\
\noalign{\smallskip}\hline
\end{longtable}
}
\normalsize

%
%
\subsection{Performance Comparison with State-of-the-art Algorithms}
We compared our method with a selection of Genetic Programming (GP) and Machine Learning (ML) based regression approaches presented in~\cite{PMLB-Regression:GECCO2018} for the ranking computed over MSE scores. In~\cite{PMLB-Regression:GECCO2018} each of the GP-based algorithms was ran for 100,000 evaluations (i.e. population size $\times$ number of generations, see details in~\cite{PMLB-Regression:GECCO2018}), except the \texttt{eplex-1m} that was allowed to run for 1 million generations. We labelled our proposal as `CFR' (standing for the fact that it is based on `Continued Fraction Regression') and it was ran 10 times to compare with the outcomes presented in ~\cite{PMLB-Regression:GECCO2018} (that also run the methods 10 times). A detailed description of the algorithms that are compared against CFR can be found in~\cite{PMLB-Regression:GECCO2018}, however, we are listing them to enhance the readability of the paper. They are:

\begin{enumerate}
    \item Genetic Programming-based Algorithms:
    \begin{enumerate}
        \item \texttt{afp}: Age-fitness Pareto Optimization~\cite{Schmidt2011:afp} 
        \item \texttt{eplex}: $\epsilon$-Lexicase selection~\cite{LaCava:2016:eplex}
        \item \texttt{eplex-1m}: Variation of \texttt{eplex} with stopping criteria of 1 million evaluations (population size $\times$ generations)~\cite{LaCava:2016:eplex}
        \item \texttt{gsgp}: Geometric Semantic Genetic Programming~\cite{Alberto:2012:gsgp}
        \item \texttt{mrgp}: Multiple Regression Genetic Programming~\cite{Arnaldo:2014:mrgp};
    \end{enumerate}
    \item Machine Learning-based Algorithms:
    \begin{enumerate}
        \item \texttt{adaboost}: Adaptive Boosting (AdaBoost) Regression (\texttt{ada-b})~\cite{Drucker:1997:ada-br}   
        \item \texttt{gradboost}: Gradient Boosting Regression (\texttt{grad-b})~\cite{Friedman00:grad-b}
        \item \texttt{kernel-ridge}: Kernel Ridge  (\texttt{krnl-r})~\cite{Murphy:2012:MLP:krnl-r}
        \item \texttt{lasso-lars}: Least-Angle Regression with Lasso  (\texttt{lasso-l})~\cite{Tibshirani94:lasso}
        \item \texttt{linear-regression}: Linear Regression  (\texttt{l-regr})~\cite{Efron04:lars}
        \item \texttt{linear-svr}: Linear Support Vector Regression  (\texttt{l-svr})~\cite{Smola2004:l-svr}
        \item \texttt{mlp}: Multilayer Perceptrons (MLPs) Regressor~\cite{kingma2014adam}
        \item \texttt{rf}: Random Forests Regression~\cite{Breiman2001}
        \item \texttt{sgd-regression}: Stochastic Gradient Descent Regression  (\texttt{sgd-r})~\cite{Pedregosa:2011:scikit}
        \item \texttt{xgboost}: Extreme Gradient Boosting  (\texttt{xg-b})~\cite{Chen:2016:xgboost}.
    \end{enumerate}
\end{enumerate}

To illustrate the differences in the performances of the algorithms, we will use \textit{violin plots}\footnote{\url{https://seaborn.pydata.org/generated/seaborn.violinplot.html}}. These plots combine the \textit{box-and-whisker plot} with the quantitative distribution of the results. The box-and-whisker plot represents the five number descriptive statistics. The ends of the box represent the \textit{upper} and \textit{lower quartiles}, a vertical line inside the box represent the \textit{median} and two whiskers outside the box extend to the \textit{highest} and \textit{lowest} value of the observations. In addition to these statistics, the colored violin shows the quantitative distribution of the results.

\subsubsection{Performance Comparison with GP-based algorithms}

\begin{figure}
\centering
\subfigure[Training Performance of Algorithms]{\includegraphics[width=0.475\columnwidth]{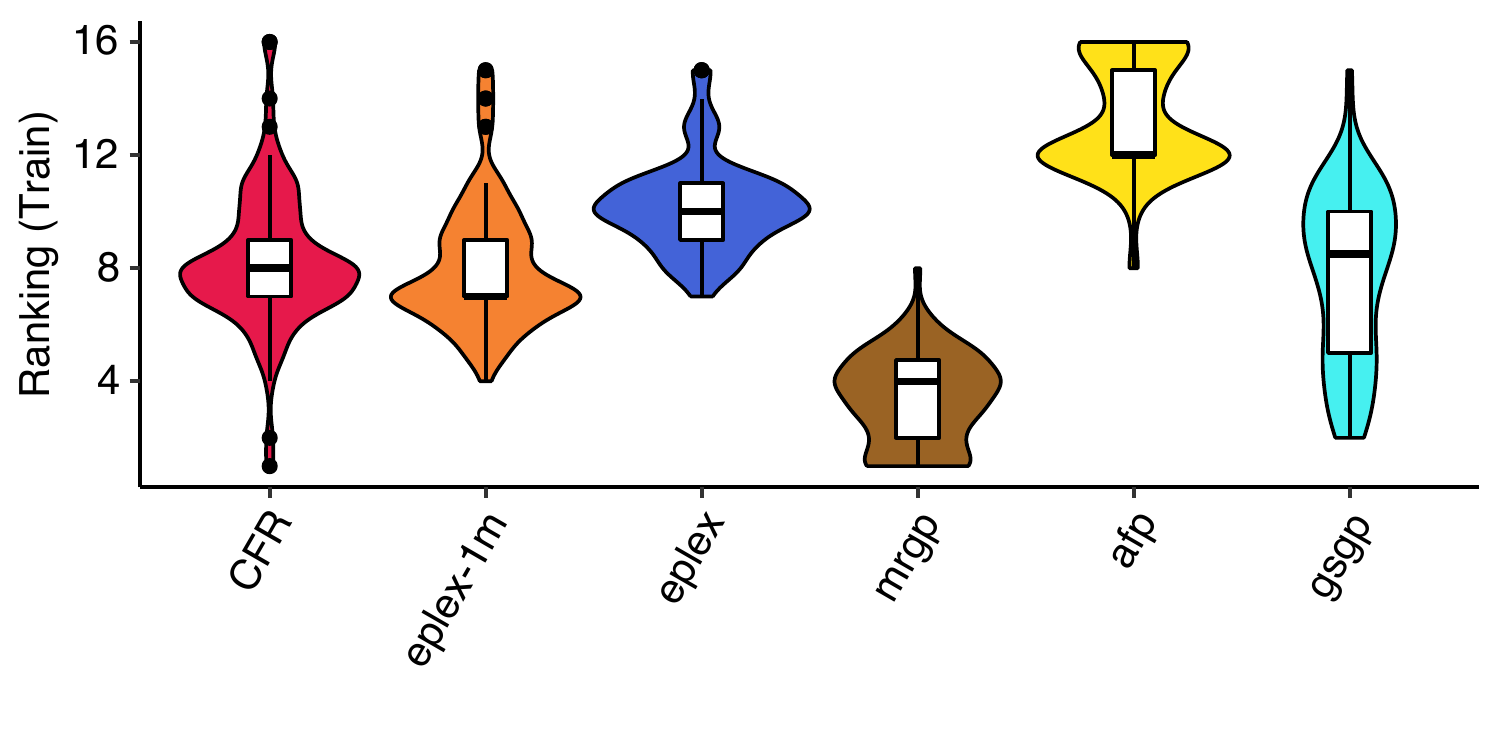}%
\label{fig_PMLB_gp_train}}
\hfil
\subfigure[Testing Performance of Algorithms]{\includegraphics[width=0.475\columnwidth]{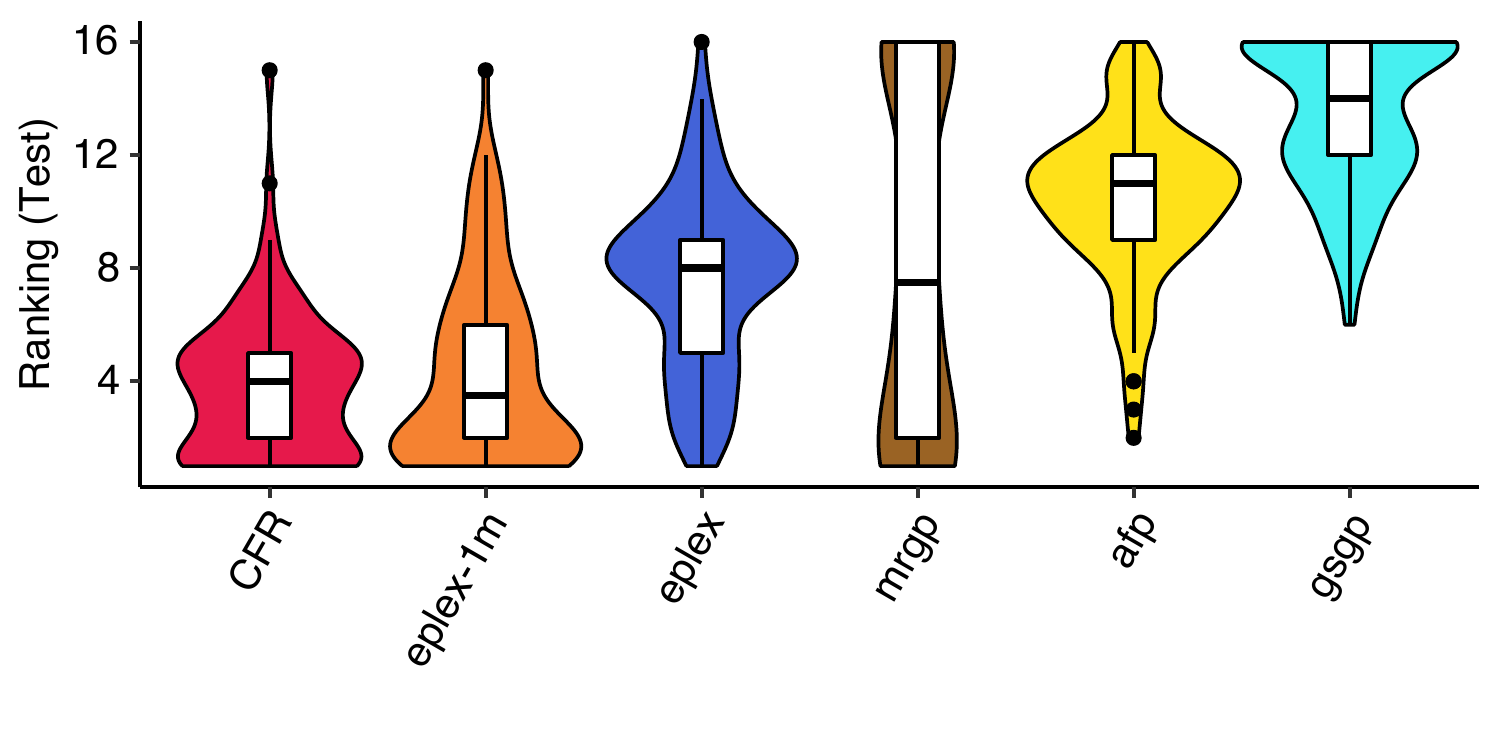}%
\label{fig_PMLB_gp_test}}
\caption{Median Ranking of the performance of CFR compared with GP-based algorithms (as reported in~\cite{PMLB-Regression:GECCO2018}) for the MSE score on a) Training and b) Testing datasets.}
\label{fig_res_PMLB_GP}
\end{figure}

Fig.~\ref{fig_res_PMLB_GP} shows the median ranking of the GP-based approaches (including the proposed CFR) for 10 repeated runs on the 94 benchmark datasets. From the performance in Fig.~\ref{fig_PMLB_gp_train} on the training set, we can observe that \texttt{mrgp} achieved the best results median ranking among six algorithms. The proposed CFR appeared the \nth{3} in median ranking for the training set. However, observing the generalization performance of those algorithms on the testing set (Fig.~\ref{fig_PMLB_gp_test}), our method has improved to \nth{2} rank.
We note now that \texttt{mrgp}, the best-performing approach in the training set, is now in the \nth{3} place for testing (and very close to \texttt{eplex}). In addition, \texttt{mrgp}'s \nth{75} percentile ranking is the \nth{2} worst performance exhibited by all GP-based methods in the testing set. Now if we consider the \nth{75} percentile of the median ranking on testing, CFR appears as the best regression method among the GP-based approaches. In terms of generalization performances, we can claim that the proposed CFR exhibited better performance than many of the state-of-the-art GP-based regression approaches.

\subsubsection{Performance Comparison with ML-based algorithms}

\begin{figure}
\centering
\subfigure[Training Performance of Algorithms]{\includegraphics[width=0.9\columnwidth]{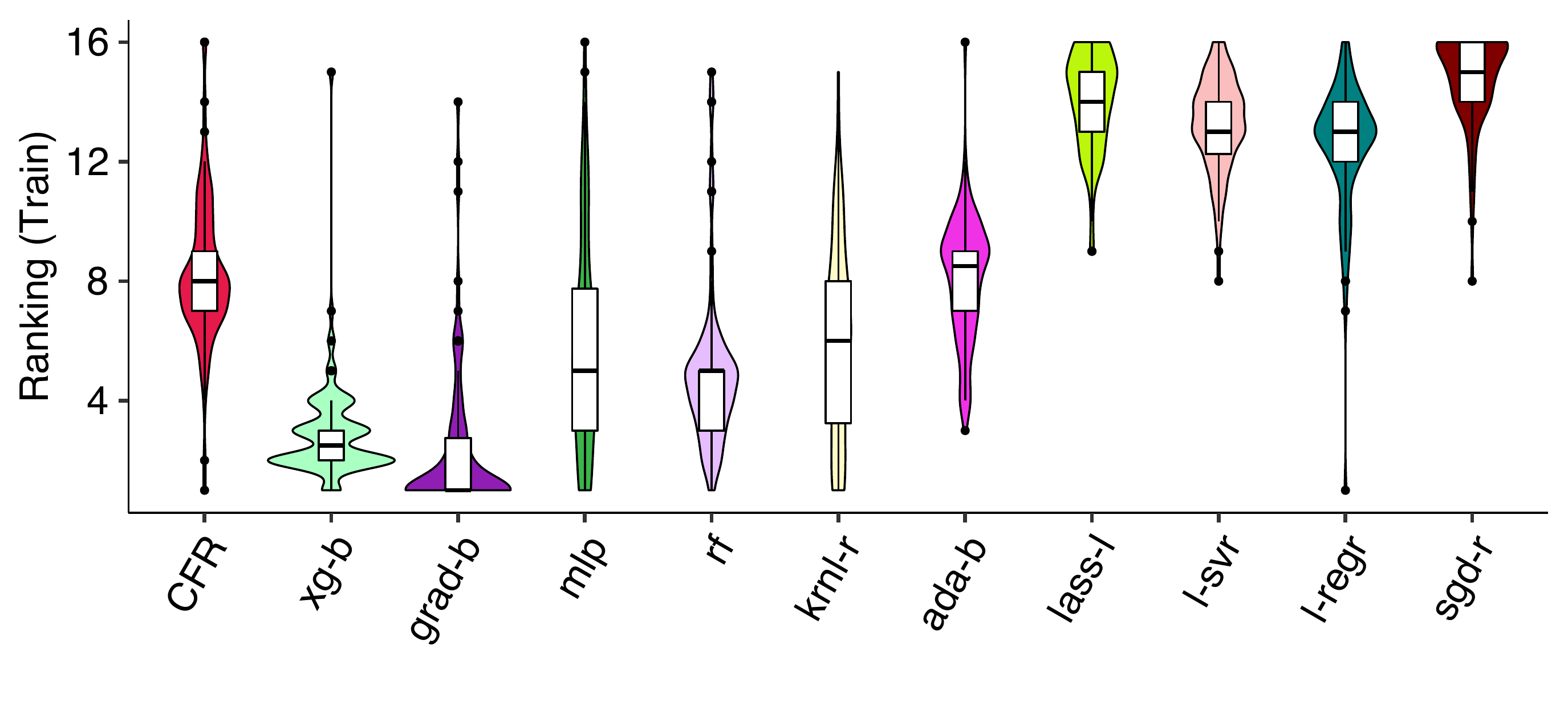}%
\label{fig_PMLB_ml_train}}
\subfigure[Testing Performance of Algorithms]{\includegraphics[width=0.9\columnwidth]{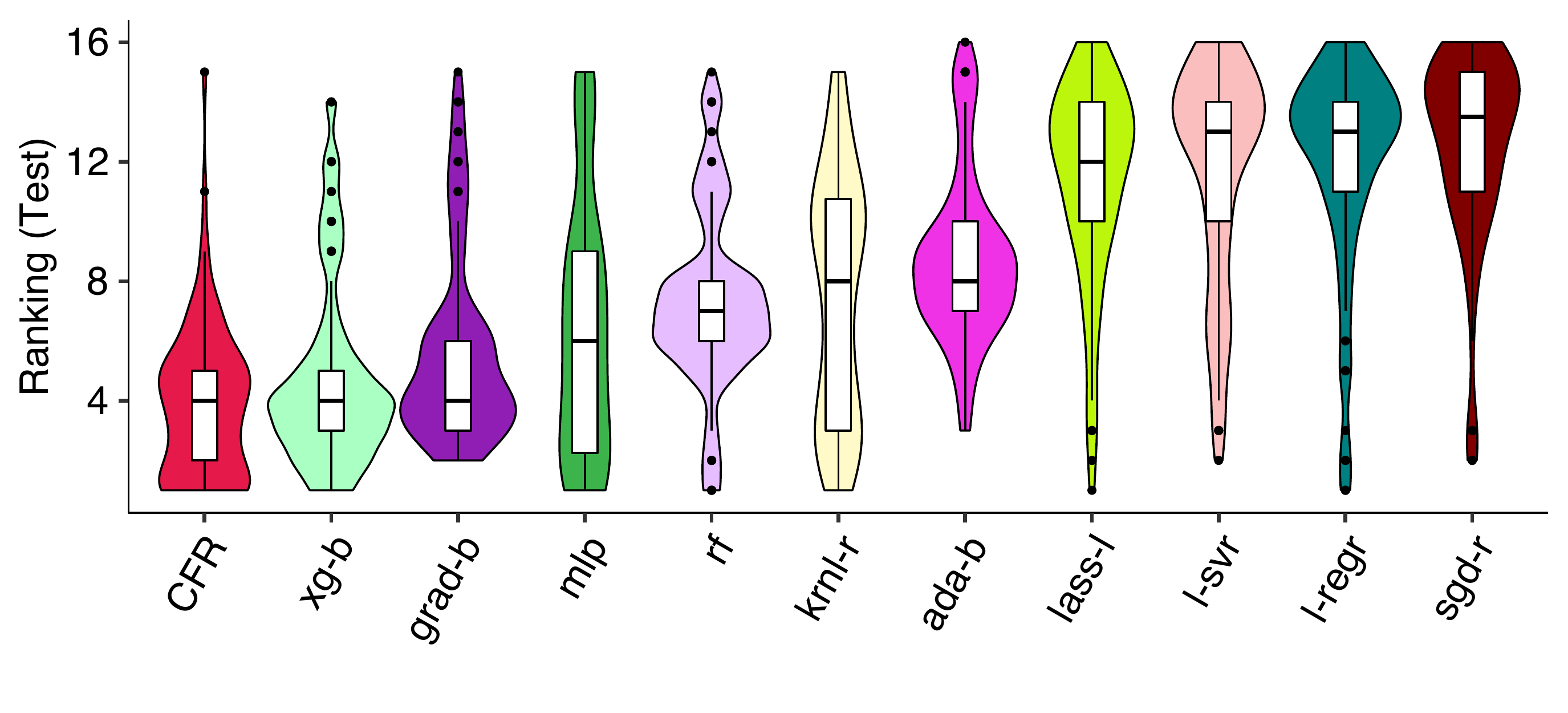}%
\label{fig_PMLB_ml_test}}
\caption{Median Ranking of the performance of CFR compared with ML-based algorithms (as reported in~\cite{PMLB-Regression:GECCO2018}) for the MSE score on a) Training and b) Testing datasets. The leftmost three are the boosting-based methods.}
\label{fig_res_PMLB_ML}
\end{figure}

Fig.~\ref{fig_res_PMLB_ML} shows the median ranking of the 10 ML-based approaches and the proposed CFR for 10 repeated runs on the 94 benchmark datasets. On the training set (Fig.~\ref{fig_PMLB_ml_train}), we can observe that the \texttt{gradboost} achieved the best middle quartile performance. The proposed CFR is, at least, better than four rightmost machine learning algorithms (\texttt{linear-svr}, \texttt{linear-regression}, \texttt{lasso-lars} and \texttt{sgd-regression}). 

When we look at the generalization performance of those Machine Learning (ML)-based algorithms on the testing set (Fig.~\ref{fig_PMLB_ml_test}), \texttt{CFR} achieved the joint top position with the \texttt{xgboost} and \texttt{gradboost} while considering the middle quartile. Moreover, based on the \nth{25} percentile result, the \texttt{CFR} achieved the \nth{1} position among all ML-based methods. However, the two best-performed boosting-based algorithms in training set, \texttt{xgboost} and \texttt{gradboost}, swapped their ranks on testing while we consider the \nth{25} percentile of the median ranking. 
We note that  \texttt{gradboost} was unable to achieve \nth{1} ranking position for any of the testing sets, despite it being ranked \nth{1} for some experiments in testing sets.

\begin{table}
 \caption{Number of times an algorithm ranked \nth{1} (1st) and Last (L) for both the train (t) and test (T) sets, ordered by descending frequency of ranked \nth{1} on the test sets ($T^{1st}$).\label{tab:freq_1st} }
 \centering
\footnotesize{
\begin{tabular}{lrrrrlrrrr}
\toprule
Algorithm & $t_{L}$ & $T_{L}$ & $t^{1st}$ & $T^{1st}$ & Algorithm & $t_{L}$ & $T_{L}$ & $t^{1st}$ & $T^{1st}$ \\
\midrule
CFR & 2 & 0 & 1 & \textbf{23} & eplex & 0 & 1 & 0 & 1 \\
mrgp & 0 & 26 & 17 & \textbf{23} & lass-l & 19 & 4 & 0 & 1 \\
eplex-1m & 0 & 0 & 0 & 19 & ada-b & 1 & 1 & 0 & 0 \\
mlp & 1 & 0 & 11 & 12 & afp & 22 & 1 & 0 & 0 \\
xg-b & 0 & 0 & 6 & 7 & grad-b & 0 & 0 & \textbf{38} & 0 \\
krnl-r & 0 & 0 & 11 & 3 & gsgp & 0 & \textbf{36} & 0 & 0 \\
rf & 0 & 0 & 2 & 3 & l-svr & 5 & 8 & 0 & 0 \\
l-regr & 3 & 6 & 1 & 2 & sgd-r & \textbf{41} & 11 & 0 & 0 \\

\bottomrule
\end{tabular}
}
\end{table}

To better understand the global picture, Table~\ref{tab:freq_1st} shows the number of times an algorithm ranked \nth{1} and last for the mean of 10 runs on both the train and test splits of 94 benchmark datasets. We have sorted the algorithms in decreasing order of the number of times it ranked \nth{1} on the testing set ($T^{1st}$). We note that there are 6 algorithms (\texttt{adaboost}, \texttt{afp}, \texttt{gradboost}, \texttt{gsgp}, \texttt{linear-svr} and \texttt{sgd-regression}) which were unable to achieve \nth{1} ranking position for any of the testing sets. Although, \texttt{gradboost} was ranked \nth{1} for the highest number of times (38 out 94 datasets) in training, it never achieved the \nth{1} rank in testing. The \texttt{sgd-regression} was the worst-performing algorithm among all by ranked last in training for the highest number (41) of datasets. Considering the generalization performance, \texttt{gsgp} was the worst among 16 algorithms for being ranked last in 36 testing set. The proposed CFR and \texttt{mrgp} achieved the joint \nth{1} position in generalization performance for being ranked \nth{1} in highest number of datasets (23 out of 94). It is also notable that, \texttt{mrgp} became last for 26 datasets out of 94 for the ranking in the testing set. Hence, CFR superseded the \texttt{mrgp} by never being last in testing performances for any of the 94 datasets.

\subsection{Statistical Significance Testing\label{sec:stat-test}}

To check whether there are any significant differences in the median rankings of the algorithms based on their MSE performances, we have used a modification of the Friedman test~\cite{friedman1937use} by Iman and Davenport~\cite{iman1980approximations}. The obtained $p$-value ($p$-value $<$ $2.2e\text{-}16$, Corrected Friedman's chi-squared = $68.089$ with df1 = 15, df2=1395) indicates that the null hypothesis \textit{``all the algorithms perform the same''} can safely be \textit{rejected}. Therefore, we proceeded with the post-hoc test.

\begin{figure}
     \centering
     \includegraphics[width=0.9\columnwidth]{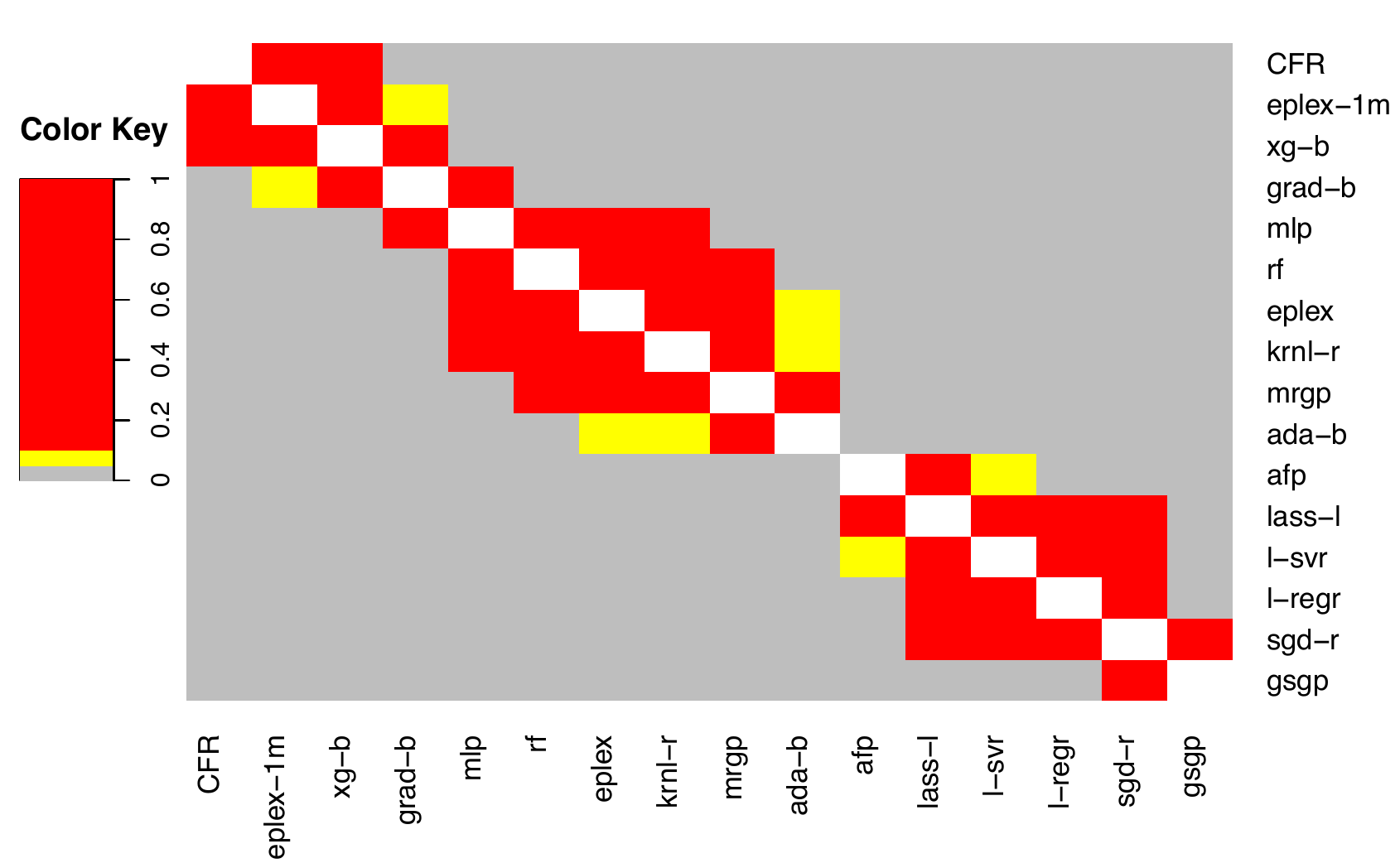}
     \caption{Heatmap showing the levels of $p$-values obtained by the Post-hoc pairwise Test using Freidman's Aligned Rank Test for the performances of the algorithms.\label{fig:heatmap}}
\end{figure}

For post-hoc test, we have applied Freidman's post-hoc test and plotted the $p$-values in the heatmap shown in Fig.~\ref{fig:heatmap}. Grey color indicates the $p$-values with `significant differences' ($p \leq 0.05$) between the pair of algorithms being tested. The yellow ($0.05 < p \leq 0.10$) and red ($ p > 0.10$) color expressed there exist `no significant' differences between the pairs of algorithms.

\begin{figure}
     \centering
     \includegraphics[width=0.9\columnwidth]{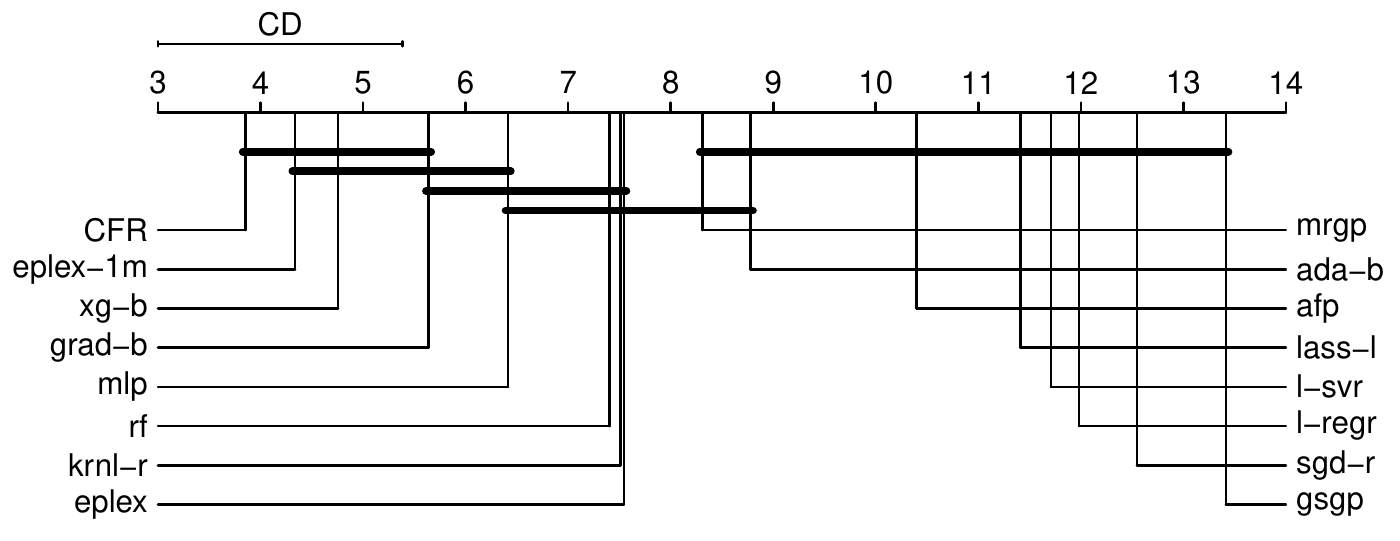}
     \caption{Comparison of all algorithms against each other with the Nemenyi test. This critical difference (CD) plot used the median rankings of algorithms in each of the 94 benchmark datasets. Groups of algorithms that are not significantly different (at $p = 0.05$) are connected.\label{fig:cd-plot}}
\end{figure}

In addition to the post-hoc results, the differences among algorithms can also be easily visualized using a Critical Difference (CD) diagram proposed in~\cite{demvsar2006statistical}, which is based on the Nyemeni post-hoc test and might have slightly different results from the pairwise Friedman test. Here, the \textit{critical difference} of rankings between the algorithm are computed and if the performance difference of any pair of algorithms are greater than the critical difference, they are regarded as ``significantly different''. The algorithms are placed on x-axis at the place of their median ranking, then statistically ``non significant'' algorithms are connected with horizontal lines. 
We have plotted the CD graph (in Fig.~\ref{fig:cd-plot}) using the implementation of~\cite{calvo2016scmamp} for our experiments with a significance threshold of $p=0.05$. For this test, the critical difference is found to be $2.3834$ and \textit{significantly different} algorithms are \textit{not connected} by the horizontal line. Form the plot it is clear that there is \textit{no significant differences} found in the rankings of 
\texttt{CFR, eplex-1m, xgboost, gradboost} algorithms on 94 benchmark datasets. Moreover, the median ranking of \texttt{CFR} is within 3 to 4, which is the best result among all of the algorithms.

\subsection{Runtime Comparison of the Algorithms}
To compare the running time requirements of CFR with GP-based methods, we used an Ubuntu 18.04 Computer with 4 CPU cores and 20 GB RAM to execute all algorithms. We also selected 10 datasets (USCrime, Fac.Salaries, autoPrice, elusage, anlc\_vehicle, anlc\_neavote, anlc\_elect.2000, sl\_ex1714, rabe\_266, chatfield\_4) for which the GP methods required the least amount of reported runtime in~\cite{PMLB-Regression:GECCO2018} to facilitate this benchmarking. We have limited the running time for each of the algorithms to 72 hours in our experimental setup. 
\begin{figure}[!htbp]
\centering
\includegraphics[width=0.9\columnwidth]{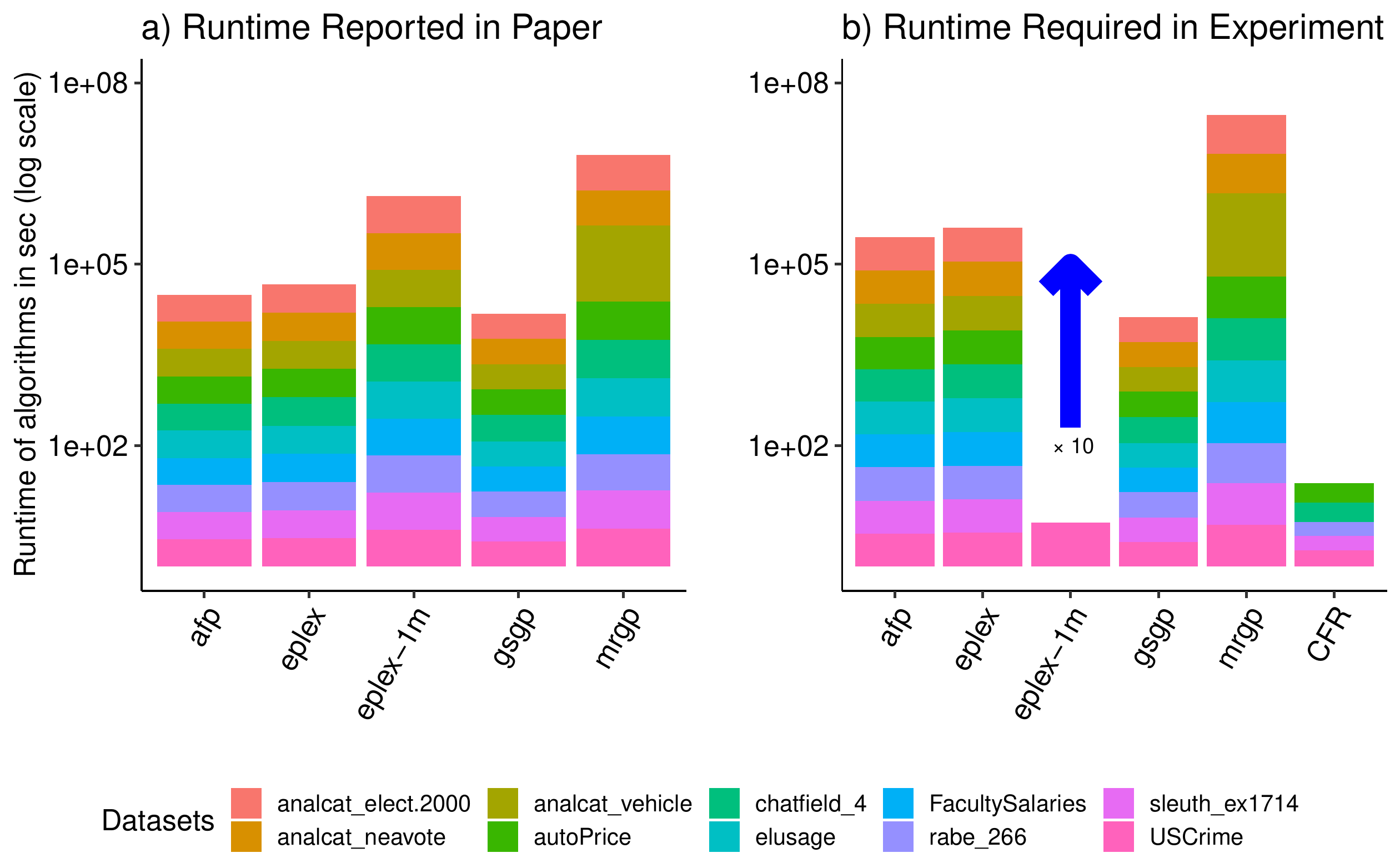}
\caption{Total running time of the algorithms in seconds (in logarithmic scale) for 10 datasets selected from the benchmark. a) Total running time reported in the paper~\cite{PMLB-Regression:GECCO2018} vs. b) Total running time required for the algorithms in the experimental computer for the 10 datasets. In the experimental computer, \texttt{eplex-1m} only completed one run in a dataset so the total time is expected to be at least ten times more than the one in the graphic (we indicated it with an arrow that shows this fact).}
\label{fig_runtime}
\end{figure}

Figure~\ref{fig_runtime} shows the total running time requirements of the algorithms for the 10 selected datasets in a) as reported~\cite{PMLB-Regression:GECCO2018} and b) experimental environment of \cite{PMLB-Regression:GECCO2018}. It is notable that the \texttt{eplex-1m} was able to finish the execution in only one of the datasets (USCrime) using
3115.36 times more CPU than CFR. Now we compare the ratio of the required running time by the reported value. 

We found that the running time of \texttt{afp} in the experimental computer is higher by a factor of $4.93$ than the reported values using the hardware of \cite{PMLB-Regression:GECCO2018}. Similarly, the runtime of \texttt{eplex} has increased by a factor of $5.17$ in the experimental computer. Besides, \texttt{gsgp} required an average CPU times by a factor of $0.93$ than the values reported in~\cite{PMLB-Regression:GECCO2018}. Finally, \texttt{mrgp} has exhibited a runtime increment by a factor of $4.18$ in the experimental environment. In the experimental setup, we have found that the \texttt{CFR} superseded the GP-based algorithms in terms of running time.

\subsection{The performance profile of all algorithms}

Dolan and Mor{\'e}~\cite{Dolan2002} proposed an approach to visualize the information that results from running multiple optimization algorithms over a set of datasets. This approach leads to the creation of a `performance profile'. We will use it here to show the results of the 16 learning algorithms over the 94 datasets studied.  

The performance profile is an $(x,y)$-plot and there are as many discontinued functions in the graph as algorithms are compared. To create it, we first need to identify, for each of the 94 datasets on which generalization has been tested, the minimum average MSE value observed (i.e. the best average MSE result obtained by one of the 16 algorithms, where the average is obtained using the 10 independent runs). This value is then taken as a useful reference; it then used to calculate the percentual relative error differences observed.
Accordingly, assume that we have the curves of two algorithms $A_1$ and $A_2$ that respectively cross the points $(500,30)$ and $(500,70)$. The $x$ value of 500 indicates that for a relative error difference corresponding to 500 percent to the minimum average MSE value observed, algorithm $A_1$ has obtained such relative error difference (on average) in only 30 percent of the 94 datasets. In comparison, the algorithm $A_2$ has achieved such a performance in 70 percent of the datasets.

\begin{figure}[!htbp]
\centering
\subfigure[All Algorithms on the $x$-axis interval from 0 to 2000]{\includegraphics[width=0.9\columnwidth]{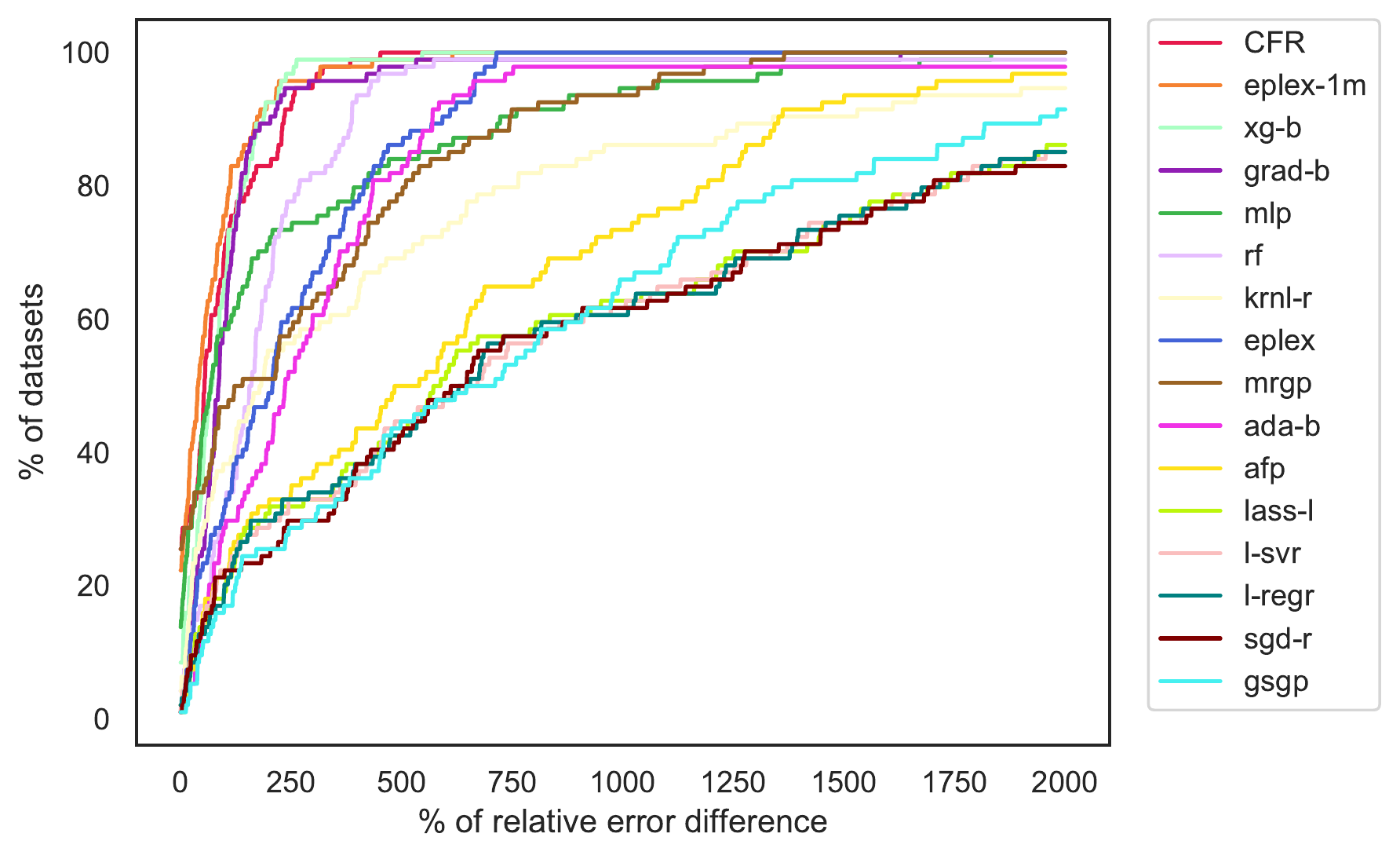}%
\label{fig_perf_prof_all}}
\subfigure[All Algorithms on the $x$-axis interval from 0 to 500]{\includegraphics[width=0.9\columnwidth]{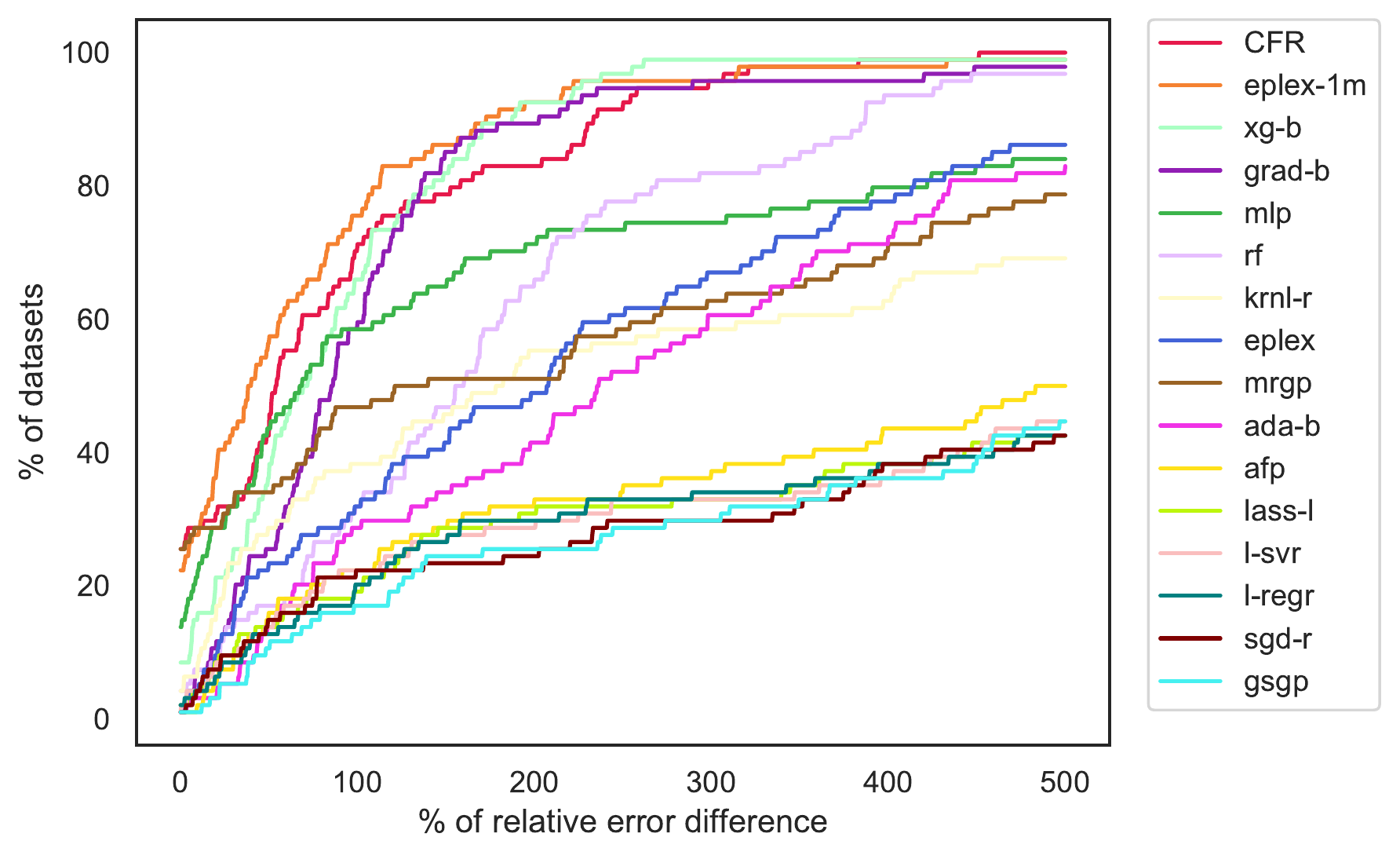}%
\label{fig_perf_prof_zoom}}
\caption{Performance Profiles for all algorithms on the $x$-axis intervals: a) [0, 2000] and b) [0, 500].}
\label{fig_perf_profiles}
\end{figure}

Varying the value of the percentual relative error difference, we can generate a curve that will go from 0 to 100 percent (for all algorithms). On a given interval on the $x$-axis, an algorithm 
$A$ would dominate an algorithm $B$ if the $y$ value of any point of its performance profile curve is larger than the of $A$ for the same value of $x$, regardless of which value of $x$ we are choosing on that interval. 
Fig.~\ref{fig_perf_profiles} shows the performance profiling of the algorithms on two different intervals which allows to have an overall picture of the performance of the whole group of algorithms tested in this work. The order of the algorithms in this figure is the one obtained from the statistical analysis and results presented in Fig.\ref{fig:cd-plot}.
Sub-fig.~\ref{fig_perf_prof_all} shows the profile for all algorithms on the interval [0,2000].
On that interval, only a few of the 16 learning algorithms achieved the top 100 percent mark. 
\texttt{CFR} is the first in achieving it and does it at a percentual relative error of 451. The \texttt{xgboost} achieves the same feat at 547 later followed by \texttt{eplex-1m} at 615. Sub-fig.~\ref{fig_perf_prof_zoom} shows the profiling on the $x$-axis interval of [0,500]. The performance profiles of \texttt{CFR}, \texttt{eplex-1m}, \texttt{xgboost}, and \texttt{xgboost}, relative to the other curves, indicate that for some datasets the other algorithms obtain present errors in generalization that are 5 times larger than the best one observed. This speaks of the relative robustness of \texttt{CFR}, \texttt{eplex-1m}, \texttt{xgboost} and \texttt{xgboost} in generalization performances. 


\subsection{Median MSE scores of Top 4 algorithms for 94 Datasets}
In this section, we tabulated the performance (measured in MSE score) of top four methods in Table~\ref{tab:penn-ml-10run-top-4}. These four methods (\texttt{CFR}, \texttt{explex-1m}, \texttt{grad-b} and \texttt{xgb-b}) has shown no significant differences of performances in Fig.\ref{fig:cd-plot}. We highlighted the best performances of \texttt{CFR} in bold in compared with other three method's MSE score.

\footnotesize{

\begin{longtable}{
    >{\raggedright \arraybackslash}p{1.75cm}
    >{\raggedleft \arraybackslash}p{1.25cm}
    >{\raggedleft \arraybackslash}p{1cm}
    >{\raggedleft \arraybackslash}p{1cm}
    >{\raggedleft \arraybackslash}p{1cm} 
    >{\raggedright\arraybackslash}p{1.75cm}
    >{\raggedleft \arraybackslash}p{1.25cm}
    >{\raggedleft \arraybackslash}p{1cm}
    >{\raggedleft \arraybackslash}p{1cm}
    >{\raggedleft \arraybackslash}p{1cm}
    }

\caption{The median value of MSE score for 10 runs obtained in Test sets of the best four algorithms. A bold-face in `CFR' column indicates it has achieved the best score among four algorithms for corresponding dataset. \label{tab:penn-ml-10run-top-4}}
\\

\hline\noalign{\smallskip}
{Dataset} & {eplex-1m} & {grad-b} & {xg-b} & {CFR} & {Dataset} & {eplex-1m} & {grad-b} & {xg-b} & {CFR}\\
\noalign{\smallskip}\hline\noalign{\smallskip}

\endhead
\noalign{\smallskip}\hline\noalign{\smallskip}
\multicolumn{10}{r}{\footnotesize Continue on the next page}
\endfoot
\endlastfoot

ESL	&	0.274	&	0.319	&	0.272	&	\textbf{0.268}	&	fri\_c3\_1000\_10	&	0.077	&	0.066	&	0.062	&	0.068	\\
SWD	&	0.390	&	0.405	&	0.408	&	0.432	&	fri\_c0\_1000\_5	&	0.042	&	0.074	&	0.071	&	0.106	\\
LEV	&	0.425	&	0.424	&	0.422	&	\textbf{0.353}	&	fri\_c3\_100\_5	&	0.250	&	0.182	&	0.093	&	0.125	\\
ERA	&	2.514	&	2.585	&	2.567	&	\textbf{2.450}	&	fri\_c1\_1000\_5	&	0.049	&	0.047	&	0.047	&	0.050	\\
USCrime	&	3.93e2 &	2.57e2 &	3.78e2	&	\textbf{2.20e2}	&	fri\_c3\_250\_5	&	0.126	&	0.110	&	0.108	&	\textbf{0.086}	\\
FacultySalry	&	4.035	&	8.071	&	4.111	&	\textbf{1.277}	&	fri\_c4\_250\_10	&	0.148	&	0.173	&	0.172	&	\textbf{0.133}	\\
vineyard	&	6.010	&	8.223	&	7.825	&	\textbf{4.222}	&	fri\_c4\_500\_50	&	0.079	&	0.141	&	0.114	&	0.169	\\
auto\_price	&	5.89e6	&	3.89e6	&	4.03e6	&	6.01e6	&	fri\_c3\_500\_5	&	0.103	&	0.085	&	0.071	&	0.078	\\
autoPrice	&	4.17e6	&	5.29e6	&	2.87e6	&	4.83e6	&	fri\_c3\_1000\_50	&	0.068	&	0.082	&	0.070	&	0.120	\\
cloud	&	0.110	&	0.208	&	0.144	&	\textbf{0.095}	&	fri\_c1\_1000\_25	&	0.057	&	0.070	&	0.067	&	0.076	\\
elusage	&	1.35e2	&	1.99e2	&	1.37e2	&	\textbf{65.797}	&	fri\_c0\_100\_10	&	0.149	&	0.319	&	0.307	&	0.226	\\
machine\_cpu	&	3.80e3	&	2.23e3	&	2.69e3	&	\textbf{2.10e3}	&	fri\_c2\_1000\_50	&	0.063	&	0.069	&	0.076	&	0.118	\\
a.vehicle	&	4.14e4	&	2.41e4	&	4.20e4	&	\textbf{1.55e4}	&	fri\_c4\_1000\_10	&	0.050	&	0.074	&	0.060	&	0.063	\\
vinnie	&	2.287	&	2.860	&	2.659	&	\textbf{1.934}	&	fri\_c0\_100\_5	&	0.152	&	0.193	&	0.164	&	0.202	\\
pm10	&	0.640	&	0.431	&	0.399	&	0.621	&	fri\_c2\_500\_50	&	0.049	&	0.101	&	0.109	&	0.130	\\
a.neavote	&	1.180	&	0.818	&	0.917	&	\textbf{0.401}	&	fri\_c2\_500\_10	&	0.064	&	0.088	&	0.097	&	0.081	\\
a.elect2000	&	4.33e7	&	3.40e8	&	7.72e8	&	\textbf{5.09e5}	&	fri\_c3\_1000\_5	&	0.059	&	0.049	&	0.048	&	\textbf{0.046}	\\
pollution	&	1.87e3	&	2.19e3	&	1.67e3	&	\textbf{1.42e3}	&	fri\_c1\_500\_5	&	0.068	&	0.077	&	0.074	&	0.075	\\
no2	&	0.272	&	0.227	&	0.210	&	0.295	&	fri\_c0\_500\_25	&	0.047	&	0.132	&	0.127	&	0.144	\\
a.apnea2	&	1.12e6	&	9.42e5	&	7.86e5	&	\textbf{6.09e5}	&	fri\_c2\_100\_10	&	0.750	&	0.321	&	0.233	&	0.320	\\
a.apnea1	&	8.16e5	&	9.98e5	&	5.28e5	&	6.96e5	&	fri\_c0\_250\_10	&	0.047	&	0.183	&	0.170	&	0.259	\\
cpu	&	1.75e2 &	2.36e3 &	8.83e2	&	\textbf{1.64e2}	&	fri\_c1\_500\_50	&	0.091	&	0.111	&	0.133	&	0.116	\\
fri\_c0\_250\_5	&	0.065	&	0.162	&	0.165	&	0.176	&	fri\_c1\_500\_10	&	0.062	&	0.086	&	0.069	&	0.075	\\
fri\_c3\_500\_25	&	0.064	&	0.090	&	0.097	&	0.081	&	fri\_c2\_500\_25	&	0.077	&	0.097	&	0.110	&	0.121	\\
fri\_c1\_500\_25	&	0.117	&	0.108	&	0.107	&	0.111	&	fri\_c4\_250\_25	&	0.157	&	0.206	&	0.178	&	0.110	\\
fri\_c1\_1000\_50	&	0.066	&	0.073	&	0.077	&	0.104	&	fri\_c3\_500\_50	&	0.101	&	0.132	&	0.131	&	0.156	\\
fri\_c4\_500\_25	&	0.127	&	0.111	&	0.095	&	0.111	&	fri\_c3\_500\_10	&	0.062	&	0.079	&	0.066	&	0.075	\\
fri\_c3\_1000\_25	&	0.056	&	0.069	&	0.059	&	0.064	&	fri\_c1\_250\_10	&	0.114	&	0.123	&	0.105	&	0.130	\\
fri\_c4\_1000\_100	&	0.047	&	0.087	&	0.072	&	0.227	&	fri\_c1\_250\_50	&	0.098	&	0.171	&	0.154	&	0.140	\\
fri\_c2\_1000\_25	&	0.051	&	0.064	&	0.070	&	0.077	&	fri\_c0\_500\_5	&	0.048	&	0.099	&	0.106	&	0.124	\\
fri\_c0\_1000\_50	&	0.044	&	0.111	&	0.113	&	0.141	&	fri\_c0\_500\_50	&	0.053	&	0.131	&	0.170	&	0.211	\\
fri\_c1\_100\_10	&	0.485	&	0.245	&	0.154	&	0.233	&	fri\_c0\_100\_25	&	0.259	&	0.419	&	0.383	&	0.365	\\
fri\_c4\_1000\_25	&	0.058	&	0.076	&	0.066	&	0.081	&	fri\_c0\_250\_25	&	0.064	&	0.209	&	0.209	&	0.224	\\
fri\_c1\_1000\_10	&	0.047	&	0.051	&	0.049	&	0.059	&	fri\_c0\_500\_10	&	0.048	&	0.153	&	0.138	&	0.202	\\
fri\_c2\_100\_5	&	0.509	&	0.277	&	0.255	&	0.259	&	fri\_c1\_100\_5	&	0.490	&	0.208	&	0.176	&	0.241	\\
fri\_c0\_1000\_10	&	0.047	&	0.096	&	0.093	&	0.155	&	fri\_c2\_250\_10	&	0.128	&	0.123	&	0.116	&	0.140	\\
fri\_c2\_250\_5	&	0.104	&	0.113	&	0.094	&	\textbf{0.086}	&	fri\_c3\_250\_25	&	0.116	&	0.194	&	0.184	&	\textbf{0.111}	\\
fri\_c2\_500\_5	&	0.079	&	0.068	&	0.073	&	\textbf{0.054}	&	sl\_ex1714	&	1.42e6	&	1.57e6	&	2.31e6	&	\textbf{6.83e5}	\\
fri\_c0\_1000\_25	&	0.039	&	0.092	&	0.093	&	0.074	&	rabe\_266	&	7.107	&	7.316	&	3.043	&	\textbf{2.643}	\\
fri\_c2\_1000\_5	&	0.046	&	0.048	&	0.045	&	\textbf{0.044}	&	sl\_case2002	&	75.772	&	56.240	&	72.363	&	\textbf{41.742}	\\
fri\_c1\_250\_5	&	0.084	&	0.103	&	0.093	&	0.093	&	rmftsa\_ladata	&	3.012	&	3.511	&	3.212	&	\textbf{2.764}	\\
fri\_c3\_250\_10	&	0.086	&	0.150	&	0.120	&	0.084	&	v.env.	&	9.624	&	9.798	&	9.540	&	\textbf{4.700}	\\
fri\_c0\_250\_50	&	0.071	&	0.238	&	0.252	&	0.254	&	sl\_ex1605	&	101.824	&	98.440	&	91.986	&	\textbf{83.998}	\\
fri\_c4\_500\_10	&	0.065	&	0.084	&	0.072	&	0.072	&	v.galaxy	&	3.13e2	&	2.68e2	&	2.24e2	&	4.34e2	\\
fri\_c2\_250\_25	&	0.179	&	0.173	&	0.156	&	\textbf{0.149}	&	chatfield\_4	&	2.82e2	&	3.84e2	&	2.88e2	&	\textbf{1.89e2}	\\
fri\_c2\_1000\_10	&	0.069	&	0.062	&	0.058	&	0.073	&	sl\_case1202	&	3.29e3 &	3.42e3	& 3.20e3	&	\textbf{1.39e3}	\\
fri\_c4\_1000\_50	&	0.051	&	0.088	&	0.068	&	0.092	&	chs\_geyser1	&	38.888	&	39.902	&	42.887	&	\textbf{31.053}	\\
\noalign{\smallskip}\hline
\end{longtable}
}
\normalsize

\section{Discussion\label{sec:discussion}}
After analyzing the median ranking of state-of-the-art regression algorithms (both of the GP and ML-based) we found that our proposed approach comparable performance while compared with all state-of-the-art approaches. The results show that the use of continued fractions is a promising new idea due to its representational power.

Moreover, the proposed algorithm, named CFR in this contribution, was ranked as \nth{1} a total of 23 times. Among fifteen algorithms, only \texttt{mrgp}) showed a similar result. However, it became last in test for the maximum number of times (26). \texttt{eplex-1m} is in the second position in terms of the number of times an algorithm become first in the ranking. Interestingly, the base algorithm \texttt{eplex} was not able to outperform any of the top 3 algorithms, including its own variant \texttt{eplex-1m}. CFR also achieves a comparable or better result at a fraction of the amount of evaluations required by \texttt{eplex-1m} and \texttt{mrgp}. Consequently, we can claim that CFR could be potentially preferred over \texttt{mrgp} due to its better overall performance and robustness in generalization capability.

The statistical test on the results (in Fig.~\ref{fig:heatmap}) unveiled the lack of significance of those differences. We have found that there are no statistically significant differences for the performance of \texttt{CFR} with the \texttt{eplex-1m} and \texttt{xgboost} algorithm. Nonetheless, running the \texttt{eplex} for a million evaluations (in \texttt{eplex-1m}) improves the performance as expected. In contrast, we note that we have executed the CFR for only 200 generations. Furthermore, the critical difference (CD) analysis (in Fig.~\ref{fig:cd-plot}) revealed that the performance of CFR is statistically comparable to the best three state-of-the-art algorithms (\texttt{eplex-1m}, \texttt{xgboost} and \texttt{grad-boost})  and better than the rest of them. Moreover, the average ranking of CFR (which was consistently between 3 and 4) for 94 datasets makes it ahead of all of the algorithms in the CD plot.

Another important issue to consider is that the CFR results have been obtained with minimal level of parameter tuning. In contrast, the hyper-parameters of the state-of-the-art algorithms in the experiment were extensively tuned using grid-search for the experiments as reported by~\cite{PMLB-Regression:GECCO2018}. Similarly, the performance of the CFR could be improved if the parameters of the algorithms are tuned per datasets or by including techniques such as \textit{boosting}. Hence, we believe there is a clear avenue for further research in this area with the joint optimization of parameters for the inclusion of boosting in a new type of memetic algorithm.

\section{Conclusion\label{sec:conclude}}
In this paper, we present a comprehensive experimental comparison of an alternative method for multivariate regression problems that uses analytic continued fractions as a representation of the mathematical models. This has led to a challenging type of nonlinear optimization problem and we have presented a memetic algorithm for this problem using a hierarchical structured population. A variant of the classical Nelder-Mead-based search was proposed as the individual optimization. We compared the performances of our proposed method with 10 machine-learning and 5 GP-based regression methods on 94 benchmark datasets. The results indicated a better generalization performance than many state-of-the-art regression approaches. In terms of the number of times an algorithm ranked \nth{1} for the generalization in the experimental setup, the CFR ranked the \nth{1} among the 16 methods, which indeed is a very promising outcome for a first implementation of this new idea. Fig.~\ref{fig_perf_profiles} shows that in comparison with the top 4 algorithms, our proposed method is the first at reaching ceiling of 100 percent making it more robust at a fraction of the computational time required to obtain the models with the best GP-based approach (\texttt{eplex-1m}) (e.g. for one dataset (US Crime) \texttt{mrgp} and \texttt{eplex-1m} was 1040 times and 3120 times slower, respectively).

The proposed approach can be extended by including adaptations that could explore parameter optimization of the memetic algorithm to improve the performance in generalization. We foresee that more sophisticated gradient descent methods could be used in place of the Nelder-Mead solver as an optimizer in our memetic algorithm. 

In terms of running time, our approach with GP-based methods, the memetic algorithm exhibited better performances compared against all of the GP-based methods. \texttt{CFR} required the least amount of CPU time to execute on 10 selected datasets. The \texttt{eplex-1m} was the most compute extensive GP-based method. However, further possibilities exist to improve the running time of our memetic algorithm including: adopting more efficient local search approach, adding the capability to adopting the local-search over distributed computing, and/or utilize more powerful processors (GPUs or TPUs). We encourage the computing community to explore the possibilities of this new approach for regression involving analytic continued fractions and to extend to other domains of machine learning and artificial intelligence (e.g. classification). 

\section*{Acknowledgements}
We thank Dr Markus Wagner, School of Computer Science at The University of Adelaide, Australia for his thoughtful comments that helped us to improve an earlier version of the manuscript. We also thank the members of Prof. Jason H. Moore's research lab at the University of Pennsylvania, USA, for making both the source code of their experiments and the Penn Machine Learning Benchmarks datasets available. M.N.H. and P.M. thank Renata Sarmet from the Universidade Federal de S\~{a}o Carlos, Sao Paulo, Brazil for discussion about the performance profile plot and sharing her Python code to produce one of the figures.

\section*{CRediT author statement}
\textbf{Pablo Moscato:} Conceptualization, Methodology, Formal analysis, Investigation, Writing - Original Draft, Writing - Review \& Editing, Supervision, Project administration, Funding acquisition. \textbf{Haoyun Sun:} Methodology, Software, Validation, Formal analysis, Investigation, Writing - Review \& Editing. \textbf{Mohammad Nazmul Haque:} Methodology, Software, Validation, Formal analysis, Investigation, Writing - Original Draft, Writing - Review \& Editing, Visualization.

\bibliographystyle{unsrt}  
\bibliography{cfrV2}

\end{document}